\documentclass[letterpaper, 10 pt, conference]{IEEEconf}
\IEEEoverridecommandlockouts
\usepackage{amsmath,amssymb,amsfonts}
\usepackage{algorithmic}
\usepackage{graphicx}
\usepackage{textcomp}
\usepackage{xcolor}
\usepackage{biblatex}
\usepackage{subcaption}
\usepackage{booktabs}
\usepackage{array}
\usepackage{pifont}
\usepackage[bookmarks=false,hidelinks]{hyperref}

\usepackage{url}

\newcommand{\cmark}{\textcolor{green}{\ding{51}}}
\newcommand{\xmark}{\textcolor{red}{\ding{55}}}

\bibliography{ref}
\def\BibTeX{{\rm B\kern-.05em{\sc i\kern-.025em b}\kern-.08em
    T\kern-.1667em\lower.7ex\hbox{E}\kern-.125emX}}
\begin{document}

\title{Octree Diffusion for Semantic  Scene Generation and Completion}

\author{Xujia Zhang, Brendan Crowe, and Christoffer Heckman%
\thanks{Authors are with the Autonomous Robotics and Perception Group at the University of Colorado Boulder, Boulder, CO 80309, USA.}%
}

\maketitle

\begin{abstract}

The completion, extension, and generation of 3D semantic scenes are an interrelated set of capabilities that are useful for robotic navigation and exploration. Existing approaches seek to decouple these problems and solve them one-off. Additionally, these approaches are often domain-specific, requiring separate models for different data distributions, e.g.\ indoor vs.\ outdoor scenes.

To unify these techniques and provide cross-domain compatibility, we develop a single framework that can perform scene completion, extension, and generation in both indoor and outdoor scenes, which we term Octree Latent Semantic Diffusion. Our approach operates directly on an efficient dual octree graph latent representation: a hierarchical, sparse, and memory-efficient occupancy structure. This technique disentangles synthesis into two stages: (i) structure diffusion, which predicts binary split signals to construct a coarse occupancy octree, and (ii) latent semantic diffusion, which generates semantic embeddings decoded by a graph VAE into voxel-level semantic labels.
To perform semantic scene completion or extension, our model leverages inference-time latent inpainting, or outpainting respectively. These inference-time methods use partial LiDAR scans or maps to condition generation, without the need for retraining or finetuning.
We demonstrate high-quality structure, coherent semantics, and robust completion from single LiDAR scans, as well as zero-shot generalization to out-of-distribution LiDAR data. 
These results indicate that completion-through-generation in a dual octree graph latent space is a practical and scalable alternative to regression-based pipelines for real-world robotic perception tasks.

Code is publicly available on 
\href{https://github.com/XUJIAZHANG2002/Octree-Latent-Diffusion-for-Semantic-3D-Scene-Generation-and-Completion}{GitHub}.

\end{abstract}


\section{Introduction}

Completion and generation of semantically rich 3D scenes is frequently required in robotics\cite{SceneSense2}, autonomous driving\cite{diffusiondrive}, and AR/VR applications\cite{dhiman2025reflecting}. Embodied agents must reason over observed and occluded geomtries, in order to facilitate safe navigation and long-term planning. To this end, scene generation and complete are keys tools for enhancing world understanding. Each domain poses a different data distribution challenge. In outdoor environments, LiDAR sensors provide accurate but sparse geometry; in indoor settings, RGB-D sensors and LiDAR offer partial yet incomplete coverage. In addition, outdoor models must generalize across varying scales, from compact rooms to large urban scenes, while remaining efficient enough for deployment in real-world systems.

While semantic scene completion (SSC) \cite{MonoScene,voxformer,SurroundOcc,vggt} attempts to solve this, such approaches often operate on only one category of scene or modality of sensor. Additionally, these methods are often completion-only, meaning scene extension would require a second model. Common approaches often rely on deterministic regression models, making them sensitive and incapable of scene generation.

Generative models, particularly diffusion models\cite{ddpm,LDM} have recently gained popularity, especially for the task of scene extension and generation\cite{semcity,octfusion,SceneSense}. Unfortunately, such models are also often constrained to a single domain: indoor scenes, outdoor scenes, 3D objects, etc., by virtue of their training data and architecture. Furthermore, many generate occupancy only, leaving semantics to be predicted by semantic segmentation models. Most importantly, the 3D representation used in many of these models is either dense or reductive. Dense tensor representations and 3D convolutions lead to models with scalability issues. Meanwhile, methods that reduce 3D representations to 2D lose spatial relationships, leading to losses in performance and interpretability.

For these reasons we seek a generative framework while using an underlying sparse 3D representation. Our approach uses a latent diffusion model over a sparse 3D tree structure that preserves spatial relationships while greatly reducing the memory required to represent 3D space. This ubiquitous representation also enables our model to generalize to different 3D data modalities with minimal changes to the architecture. Importantly, the generative nature of diffusion models affords us the capability not only to unconditionally generate 3D geometries and semantics, but also to perform semantics scene completion and extension via a completion-through-generation framework.

Our contributions can be summarized as follows:
\begin{enumerate}
    \item \textbf{Two-stage generative framework.} We disentangle scene synthesis into coarse structure diffusion and latent semantic diffusion, ensuring that geometry is first established before fine-grained semantics are generated.
    \item \textbf{Dual octree graph patch latent representation.} We introduce a compact yet spatially local latent space based on a dual octree graph and patch-based VAE, enabling scalable generation of large indoor and outdoor scenes with voxel-level semantics.
    \item \textbf{Unified generation and completion in multiple domains.} To the best of our knowledge, this is the first diffusion-based framework that unifies 3D generation, extension, and semantic completion in both indoor and outdoor scenes.
\end{enumerate}


\section{Related Work}

\begin{figure*}[t]
    \centering
    \includegraphics[width=\textwidth]{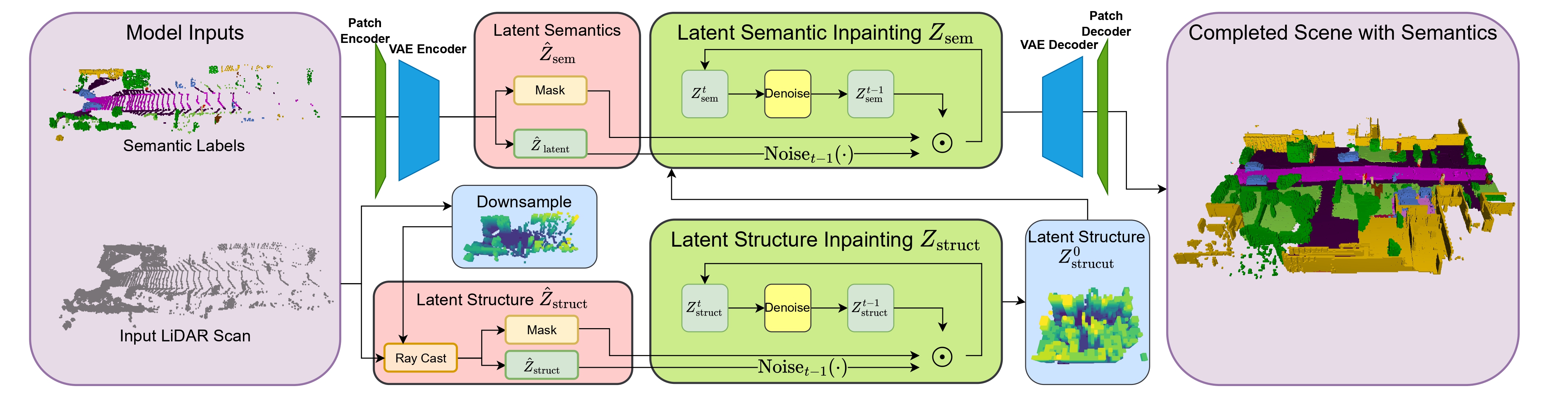}
    \caption{Our method unifies unconditional generation and LiDAR-conditioned completion within the same framework. When available, a LiDAR scan can be voxelized and used to initialize the latent structure; partial semantic voxels, if provided, are encoded and used to anchor node latents. Both structure and semantics are then introduced during postconditioned diffusion sampling, which preserves observed regions while freely synthesizing unobserved areas. When no LiDAR or semantic input is provided, the masks default to all zeros, causing the model to perform unconditional generation and sample entire scenes from pure noise. }
    \label{fig:network}
\end{figure*}

\subsection{3D Scene Generation and Completion}
Recent work in semantic scene completion (SSC) can be broadly divided into two families. End-to-end regression models, and conditioned generative models.

The regression model approach directly predicts voxel occupancies or semantics from RGB images or multi-view inputs \cite{MonoScene,voxformer}, often achieving strong accuracy with efficient inference. Their reliance on dense imagery and calibrated poses makes them less suitable for robotics in low-light or unstructured environments. Furthermore, the tendency of such discriminative methods to overfit makes them sensitive to viewpoint, sensor specifications, or other perturbations that shift the data distribution.

In contrast, generative methods treat completion as conditional generation, learning a distribution of plausible 3D scenes, which can then be sampled from to fill unobserved regions and outpaint beyond the field of view. Such models provide richer priors for completion and also serve as standalone generators when no observations are available. 

Our work follows this latter paradigm and, importantly, accepts arbitrary point clouds or occupancy grids as inputs, decoupling performance from a specific sensor.

\subsection{3D Generation via Diffusion Models}
Diffusion models have emerged as powerful generative frameworks across modalities. 
The denoising diffusion probabilistic model (DDPM)~\cite{ddpm} and its latent-space variant (LDM)~\cite{LDM} laid the foundations for scalable image synthesis. 
Extensions into 3D include voxel- and point-based diffusion for shape generation, as well as implicit radiance-field or Signed Distance Fields (SDF)-based \cite{DiffusionSDF} pipelines. 

Volumetric CNNs generalize 2D CNNs to 3D voxel grids and achieved promising results in 
3D generation~\cite{Vox2016generative}.
However, their cubic computational and memory complexity severely limits scalability to high-resolution or large-scale 3D scenes. Therefore, a central trend has been to pair diffusion with efficient 3D representations to overcome the prohibitive cost of dense volumetric modeling. 
For example, triplane features reduce volumetric data into three orthogonal 2D planes, enabling compact yet expressive volumetric generation. 
In parallel, octree-based approaches\cite{octfusion} preserve 3D spatial locality by operating directly on sparse, hierarchical structures, 
allowing diffusion to scale to high-resolution multiscale shapes. 
Together, these advances illustrate the shift towards combining diffusion priors with efficient 3D structures to enable practical scene-level generation.

\subsection{Pre- vs.\ Post-Conditioned Diffusion}
While unconditional diffusion enables generation, completion requires injecting partial observations (e.g., sparse LiDAR scans) 
into the generative process. Two main paradigms exist \cite{LatentInpaint}:  

\begin{itemize}
\item[a)] \textbf{Preconditioned inpainting} trains a model explicitly on masked inputs, learning $p(x| y)$ where $y$ denotes the observed region. 
This strategy provides high fidelity but requires retraining for each domain or mask distribution, which is costly given the variability 
of LiDAR viewpoints and sparsity patterns \cite{Diffssc}. Furthermore, they demand extensive domain-specific training data covering all possible partial inputs, and requires a sophisticated condition injection mechanism designed for a particular domain,
making adaptation across varying conditions (e.g., LiDAR
scans from different perspectives) cumbersome
and computationally expensive. 

\item[b)] \textbf{Postconditioned inpainting} instead reuses an unconditional diffusion model at inference. 
Observed regions are preserved by blending them into the denoising trajectory at each step 
\cite{BlendedLatentDiffusion,repaint}. This approach is training-free and flexible, but its effectiveness depends critically on the latent representation: 
it must preserve spatial locality to align masks precisely \cite{BlendedLatentDiffusion}.  
\end{itemize}

In principle, postconditioning is attractive for robotics since it avoids retraining under every sensor and data domain configuration. However, dense 3D latents are prohibitively large, and triplane features, while compact, collapse the volumetric structure 
onto 2D planes, making 3D masks ambiguous. This conflict motivates octree-based hierarchical representations, which remain compact yet preserve relative spatial locality, enabling accurate mask conditioning and efficient postconditioned completion at scale.

\subsection{Octree- and Graph-Based 3D Representations}
\begin{figure}[t]
    \centering
    \includegraphics[width=\columnwidth]{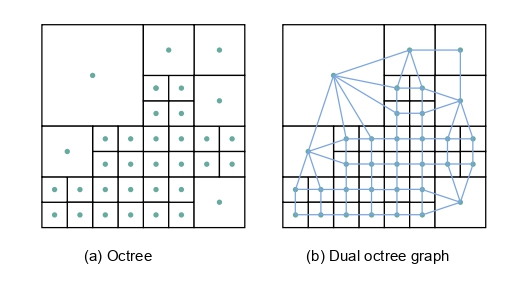}
    \caption{\textbf{Left:} A 2D rendering of an octree. \textbf{Right:} Corresponding dual octree graph.}
    \label{fig:octree}
\end{figure}
Hierarchical sparse representations like octrees provide efficient and scalable 3D modeling by focusing computation and memory on occupied regions. 
Octomap~\cite{octomap} pioneered octree-based scene mapping in robotics. 
OctNet~\cite{Riegler2016OctNetLD} demonstrated that octrees can support deep convolutional architectures, enabling high-resolution volumetric learning with manageable cost. 
O-CNN~\cite{ocnn} further optimized GPU-resident octree structures and showed their effectiveness for 3D classification and segmentation.

Later, a ``dual" representation of octrees \cite{LEON2008393} was proposed to create a semi-regular graph over arbitrary octree inputs. Building on this idea, Dual Octree Graph Networks~\cite{graphdualoctree} reorganize features into a graph of face-adjacent octree nodes (dual octree graph), enabling structured graph convolutions across scales. 
This design improves efficiency and expressiveness by capturing both local adjacency and hierarchical depth relationships.

Most recently, OctFusion~\cite{octfusion} integrated octree-based latent representations with multiscale diffusion, showing that a unified U-Net backbone can generate high-resolution, compact 3D shapes within seconds. 
This work highlighted the power of combining diffusion with hierarchical sparse structures for efficient and high-fidelity 3D generation.

Extending such paradigms from object-level shape generation to scene-level semantic generation and completion introduces new challenges. 
First, occupancy distributions differ: objects are compact and centered, whereas scenes are sparse and horizontally extended. 
Second, real-world robotics tasks require semantic prediction in addition to geometry. 
Third, completion with partial observations, which is essential for LiDAR perception, was not considered in prior octree diffusion work. 
These gaps motivate our approach, which unifies coarse occupancy diffusion, dual octree graph latent diffusion, and VAE semantic decoding, together with postconditioned blending to support LiDAR-driven inpainting and outpainting.

\subsection{Summary and Motivation}
To summarize, regression-based pipelines offer speed but lack generative flexibility, while diffusion-based models 
provide strong priors but demand efficient 3D representations. 
Recent trends show the effectiveness of triplane for scaling diffusion in 3D, and of postconditioned 
inpainting for training-free completion \cite{semcity}.
Yet no prior work has unified these advances into a framework that is (i) \textbf{two-stage}, disentangling coarse structural 
generation from semantic latent synthesis, (ii) \textbf{octree-based}, preserving locality with sparse efficiency, and 
(iii) \textbf{training-free-conditioning}, enabling LiDAR-driven completion under varied conditions, including darkness. 
Our approach fills this gap by combining coarse occupancy diffusion with dual octree graph latent diffusion and VAE decoding, 
while leveraging postconditioned blending to integrate partial observations seamlessly.

\begin{figure*}[t]
    \centering
    \includegraphics[width=\textwidth]{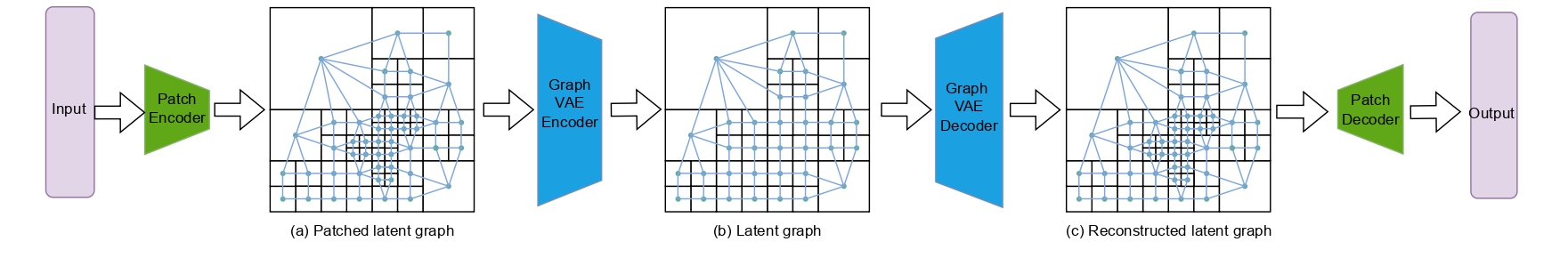}
    \caption{The Patch-VAE pipeline. A semantic voxel representation of the scene is given as input. Next, the patch encoder does spatial compression over every non-empty patch in the semantic voxel map and forms an octree in the compressed space. Then, it is converted into a dual octree graph. The VAE encoder outputs a latent representation at a shallower depth graph. The VAE decoder utilizes a shared MLP head to predict the split signal to each node at the finest depth, and reconstructs the latent graph. Finally, the patch decoder converts the latent into a semantic voxel map.}
    \label{fig:vae-pipeline}
\end{figure*}

\section{Methods}

We design a diffusion-based framework that unifies structure prediction, semantic synthesis, and training-free conditioning. Our approach is guided by three principles: (i) \textbf{sparse 3D representation}, achieved through a dual octree graph that concentrates computation on occupied regions while preserving spatial relationships; (ii) \textbf{semantic compactness}, enabled by a patch-based VAE that prevents semantic latents from being diluted by unoccupied space; and (iii) \textbf{flexible conditioning}, realized via postconditioned blended latent diffusion that integrates partial point cloud observations without retraining. The overall pipeline proceeds in two stages: first, a diffusion model predicts the coarse occupancy structure of the scene; second, latent diffusion refines per-node semantics, which are decoded into voxel labels through the VAE decoder. This design disentangles structure from semantics, supports scalable scene generation, and enables training-free completion in diverse sensing conditions. An overview of the pipeline is shown in Figure \ref{fig:vae-pipeline}.

\subsection{Dual Octree Graph Representation}

 Octrees, a common data structure used in robotics, uses hierarchical subdivisions of occupied space only, taking advantage of the natural sparsity of 3d scenes. However, octrees alone lack regular neighborhood structure, making standard convolutions intractable. We therefore convert octree leaves into a dual octree graph, where face-adjacent nodes across depths are connected (see Figure \ref{fig:octree} for architectural details) \cite{LEON2008393,dualoctree}. This dual graph forms a semi-regular structure with a finite set of edge types, enabling efficient graph convolutions, with orientation-specific kernels shared across all nodes. Dual octree graph CNNs can thus aggregate multi-scale context while maintaining sparsity. Downsampling and upsampling are implemented via grouping and splitting sibling nodes, ensuring consistent multi-resolution processing \cite{dualoctree}.

\subsection{Patch-Based Variational Autoencoder}
\begin{figure}[t]
    \centering
    \includegraphics[width=\columnwidth]{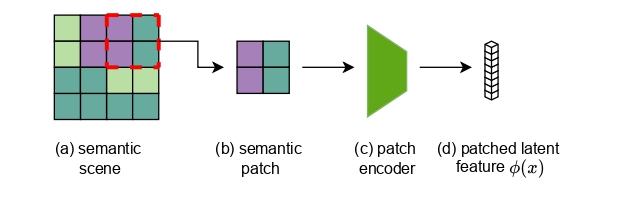}
    \caption{Given a prespecified patch size, a shared patch encoder operates on every single patch from the semantic scene, and output a latent vector for each patch.}
    \label{fig:patch_encoder}
\end{figure}

Unlike 3D shapes, real-world scenes often exhibit occupancy distribution ill conditioned for octree construction: geometry is concentrated near floors and walls,  while large volumes remain empty. Moreover, the overall geometry is often quite flat, especially in outdoors (e.g. 256*256*32). Encoding such data directly results in wasted computation, ill-conditioned octree geometries and unbalanced information flow.

We therefore introduce a patch-based VAE, which compresses local cuboid patches into compact latent vectors while discarding empty regions. More specifically, each voxel patch is processed by a shared encoder composed of convolutions followed by an MLP, yielding a latent vector (Figure \ref{fig:patch_encoder}). Patches consisting only of empty voxels are skipped, such that the latents remain as sparse as possible. The resulting octree is therefore, of shallower depth, denser, and more geometry-alligned for dual octree graph network.

The latent vectors are then embedded into a dual octree graph, where a GraphVAE aggregates multi-scale context. The encoder compresses the latent graph to a shallower depth, while the decoder expands it back to its original depth, i.e., per-patch latents. During decoding, before each deconvolution step, a shared MLP head predict the split signal for each node at the depth $d$ (split signal is described in section \ref{split-signal}). Deconvolution only operates on nodes that are split. Finally, each latent is decoded by a shared deconvolutional head into voxel-wise semantic probabilities.

 The VAE is optimized with three terms:
\begin{equation}
\mathcal{L}_{\text{VAE}} = 
\mathcal{L}_{\text{sem}} +
\mathcal{L}_{\text{octree}} +
\beta \,\mathcal{L}_{\text{KL}},
\end{equation}
where $\mathcal{L}_{\text{sem}}$ is cross-entropy for semantic reconstruction, 
$\mathcal{L}_{\text{octree}}$ is binary cross-entropy on split signals across octree levels, 
and $\mathcal{L}_{\text{KL}}$ is a KL penalty with weight $\beta$. This encourages the VAE to produce structure-aware, generative-ready latents.

\subsection{Two-Stage Diffusion for Structure and Semantics}
We disentangle generation into two diffusion stages: structure diffusion and latent diffusion.

\textbf{Structure diffusion}\label{split-signal}  predicts binary split signals of the octree, progressively constructing the occupancy structure. We model each node as either subdivided or terminal, recursively generating a dual octree graph~\cite{octfusion}. This ensures global topology is established before semantic details are added. To be more specific, beginning with a dense voxel grid of dimensions $2^D \times 2^D \times 2^D$, where each voxel holds a binary split signal (0 or 1) predicted by the structure diffusion model. Voxels labeled ``1'' are recursively subdivided into eight octants, and a dual octree graph at depth $D{+}1$ can be constructed with face-adjacent relations. Repeating this process by assigning 0/1 to nodes and subdividing those labeled ``1'' yields a dual octree graph at depth $D{+}2$. Unlike previous methods \cite{octfusion}, which trains a unified network across multiple depths, our patch-based encoder design allows us to operate at shallower depths, reducing memory and computation.

\textbf{Latent semantic diffusion} operates on the generated structure (i.e., a dual octree graph), and predicts latent codes for each octree node. Here, a dual octree graph U-Nets serve as the backbone of the diffusion model, propagating multi-scale context to produce coherent semantic latents on each node. The semantic latents are then decoded via the graph VAE decoder, and then reconstructed by the patch decoder, yielding a full voxel-level semantic scene. This factorization, structure first, semantics second, ensures valid and sparse geometry while capturing semantic diversity.

\subsection{Completion via Postconditioned Blending}
To extend generation into completion, we incorporate partial observations as conditions during sampling using postconditioned blended diffusion. 

During structure diffusion, known occupancies from LiDAR scans are forward-noised into the sampling trajectory. At each denoising step, the model’s prediction is blended with the noised reference according to a binary mask, preserving observed voxels while freely synthesizing unobserved regions. This enables both inpainting and outpainting without retraining. At each reverse step, the model’s denoised prediction $\tilde{x}_{t-1}$ is interpolated with a noised reference input $x^{\text{ref}}_{t-1}$:
\begin{equation}
    \hat{x}_{t-1} \;=\; (1 - m)\,\odot\, \tilde{x}_{t-1} \;+\; m\,\odot\, x^{\text{ref}}_{t-1},
\end{equation}
where $\odot$ denotes the element-wise product. At each denoised step, with a binary mask $m$, the entire regions are freely synthesized, while observed voxels are copied exactly from $x^{\text{ref}}_{t-1}$. Thus, the conditioned reference will gradually be injected into the denoising process, and will be kept intact at the sampling result.  

 Once the structure is generated, partial semantic information can be injected at the latent level. Known latents are assigned to observed nodes in the dual octree graph and preserved during denoising. This anchors the semantics to known regions while allowing generative refinement elsewhere. To our best knowledge, blended diffusion has not been implemented in graph networks. Our empirical results shows it does work in this semi-regular graph.

 Together, these steps form a training-free pipeline: LiDAR scans are first converted into partial occupancy masks, which guide structure diffusion; the resulting octree is then augmented with partial semantic latents, guiding latent diffusion. The final output is a completed semantic scene that is both structure-consistent and observation-faithful.

\begin{table}[h!]
\centering
\begin{tabular}{lcccc}
\hline
\textbf{Feature} & \textbf{SceneSense} & \textbf{Octfusion} & \textbf{SemCity} & \textbf{Ours}\\
\hline
Indoor & \cmark & \xmark & \xmark & \cmark \\
Outdoor & \xmark & \xmark & \cmark & \cmark \\
Objects & \xmark & \cmark & \xmark & \xmark \\
Sparse representation & \xmark  & \cmark & \cmark & \cmark\\
Scene Extension & \cmark  & \xmark  & \cmark & \cmark  \\
Scene Completion &\cmark & \cmark & \xmark & \cmark \\
Scene Generation & \cmark & \xmark & \cmark & \cmark \\
Semantics & \xmark & \xmark & \cmark & \cmark \\
\hline
\end{tabular}
\caption{Comparison of the capabilities of SceneSense, Octfusion, SemCity, and our approach.}
\label{tab:comp-of-feats}
\end{table}

\section{Experiments}

We evaluate our dual octree graph latent diffusion model on indoor and outdoor scene generation, completion, and extension tasks. 
For outdoor scenes, we train on the SemanticKITTI~\cite{Semkitti} training set and report results on the validation set. 
For indoor scenes, we use the Replica dataset~\cite{replica}. Since Replica does not provide an official train/validation split, we randomly partition the scenes into a 90/10 split and keep this split fixed across all experiments. 

SemanticKITTI contains 20 semantic categories, whereas Replica contains 92. 
Each voxel in SemanticKITTI corresponds to a $0.2\,\mathrm{m}$ cube, while in Replica each voxel represents $0.05\,\mathrm{m}$. 
We adopt a voxel grid of size $256{\times}256{\times}32$ for outdoor experiments and $128{\times}128{\times}32$ for indoor experiments. 
Patch sizes are set to $1{\times}4{\times}4$ for SemanticKITTI and $1{\times}2{\times}2$ for Replica to balance resolution with memory efficiency.All experiments were primarily conducted on NVIDIA GeForce RTX 4070 Ti and RTX 4060 Ti GPUs (16GB and 8GB VRAM, respectively).

\begin{figure}[t] 
    \centering
\begin{subfigure}[b]{0.48\columnwidth} 
    \centering
    \includegraphics[width=\linewidth]{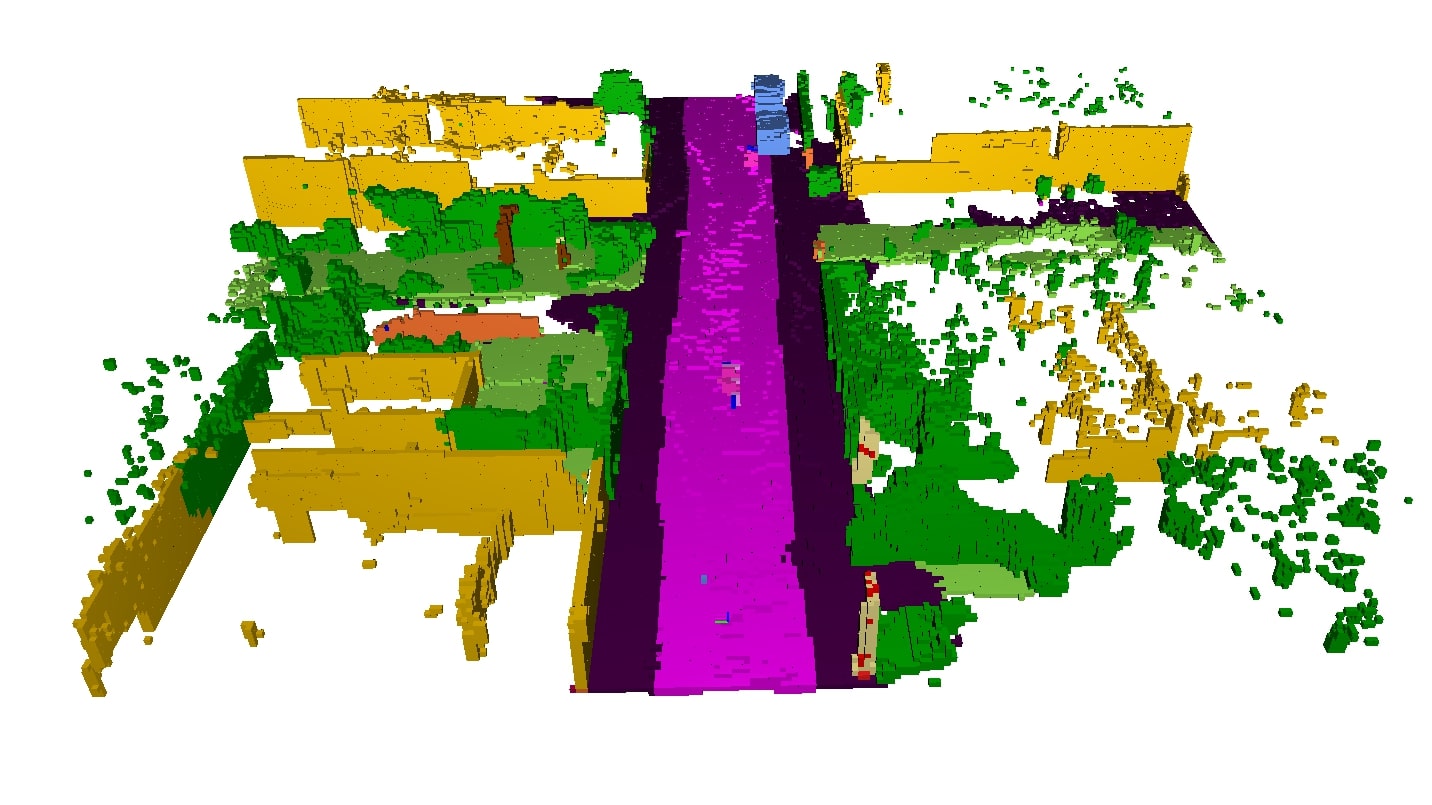}
\end{subfigure}
\hfill
\begin{subfigure}[b]{0.48\columnwidth} 
    \centering
    \includegraphics[width=\linewidth]{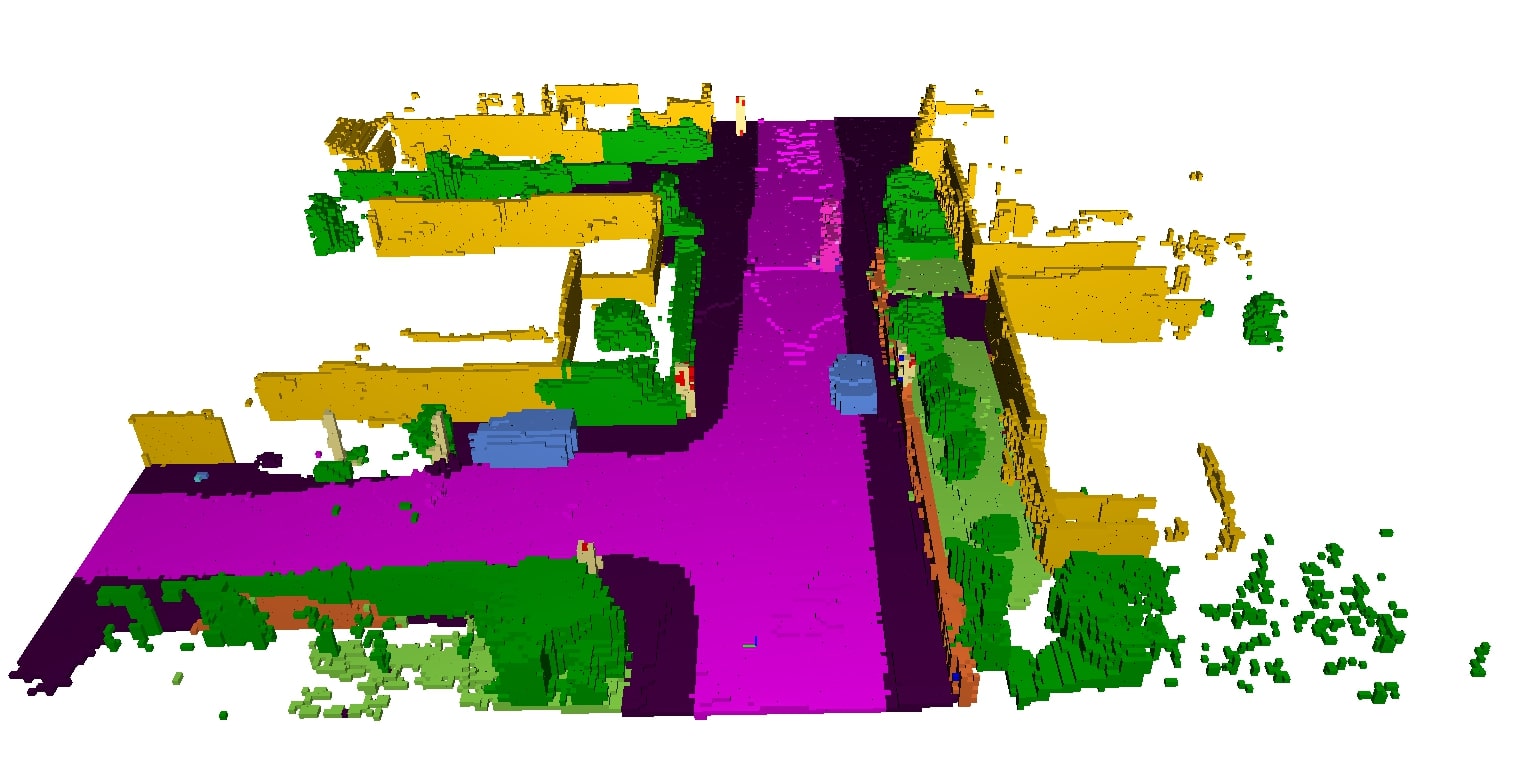}
\end{subfigure}
    

    \vskip\baselineskip
   
        \noindent
        \includegraphics[height=0.12\textheight]{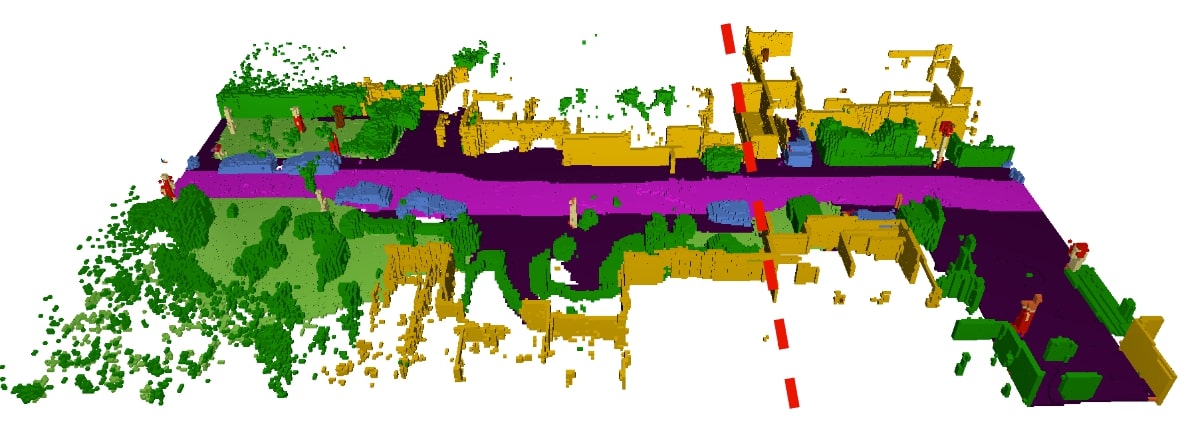}

    \caption{Scene generation results via our two-stage pipeline.
    \textbf{Top:} Two example semantic scene generations
    \textbf{Bottom:} Example semantic scene extension. Left of the dotted line is an input semantic scene, to the right is an extension of the scene via outpainting.
    }
    \label{fig:outdoor-scene-generation}
\end{figure}

\begin{figure}[t]
\centering
\setlength{\tabcolsep}{1pt} 
\renewcommand{\arraystretch}{0.8} 
\captionsetup[subfigure]{labelformat=empty} 

\begin{tabular}{@{}c@{\hskip 2pt}!{\vrule width 0.4pt}@{\hskip 2pt}c@{}}
    \begin{subfigure}[b]{0.46\columnwidth}
        \includegraphics[width=\linewidth,height=0.1\textheight]{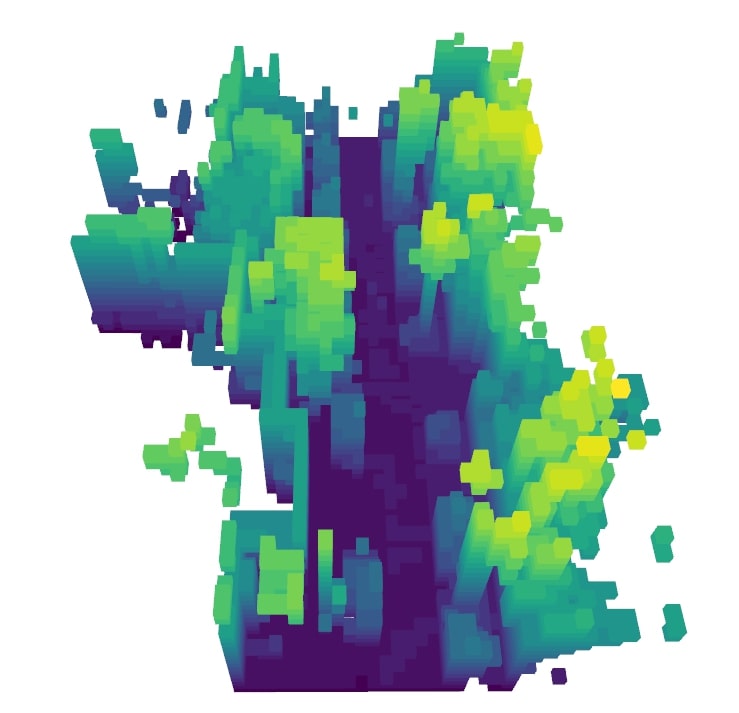}
    \end{subfigure} &
    \begin{subfigure}[b]{0.46\columnwidth}
        \includegraphics[width=\linewidth,height=0.1\textheight]{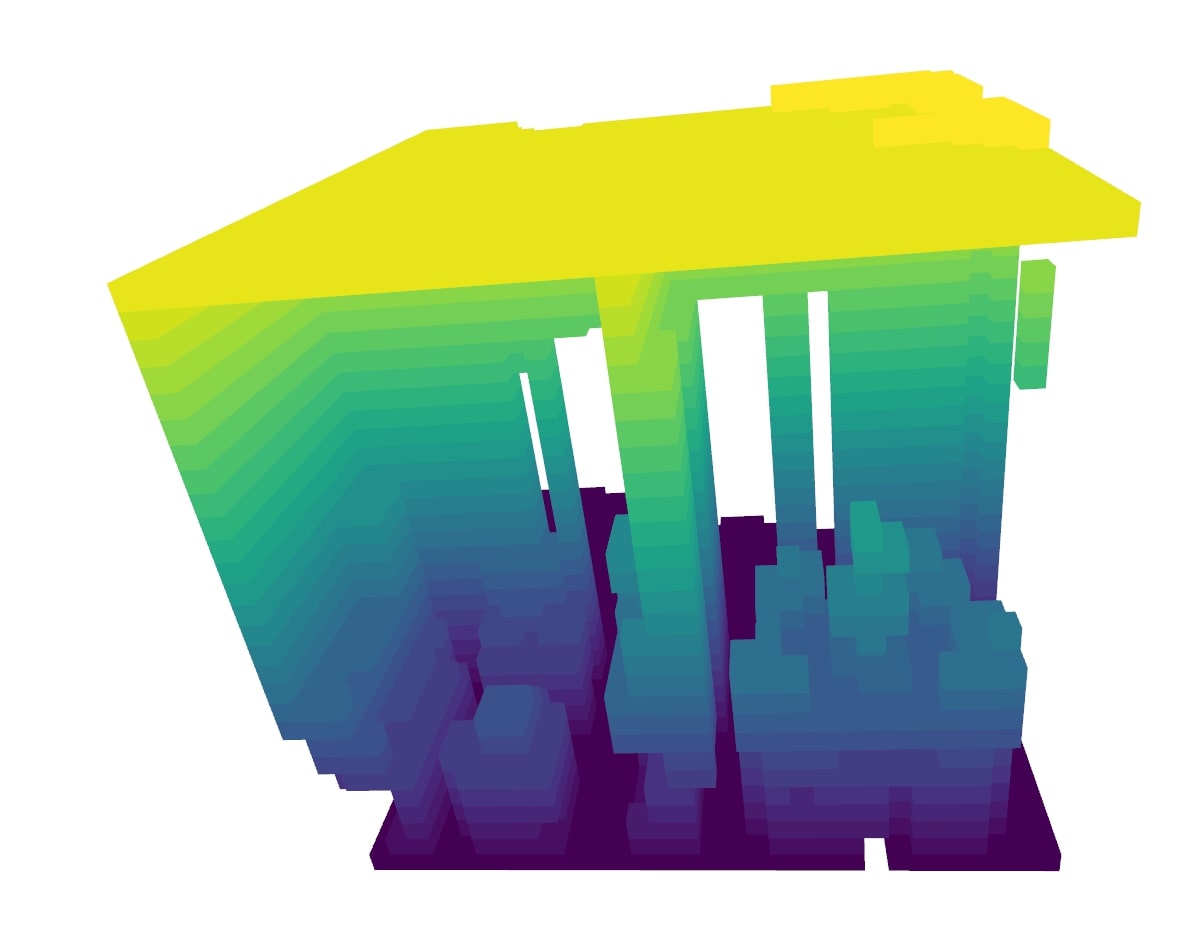}
    \end{subfigure} \\[-2pt]

    \begin{subfigure}[b]{0.46\columnwidth}
        \includegraphics[width=\linewidth,height=0.1\textheight]{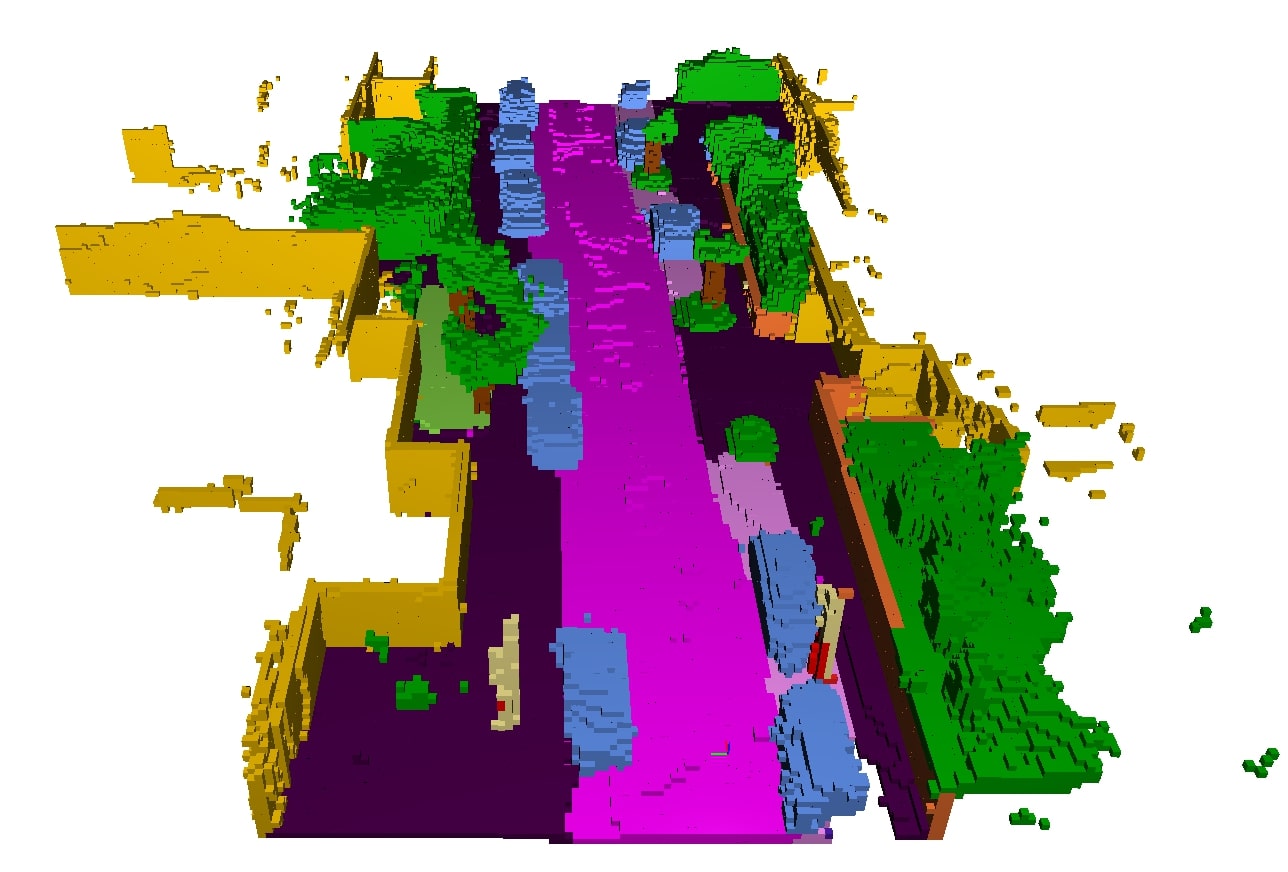}
    \end{subfigure} &
    \begin{subfigure}[b]{0.46\columnwidth}
        \includegraphics[width=\linewidth,height=0.1\textheight]{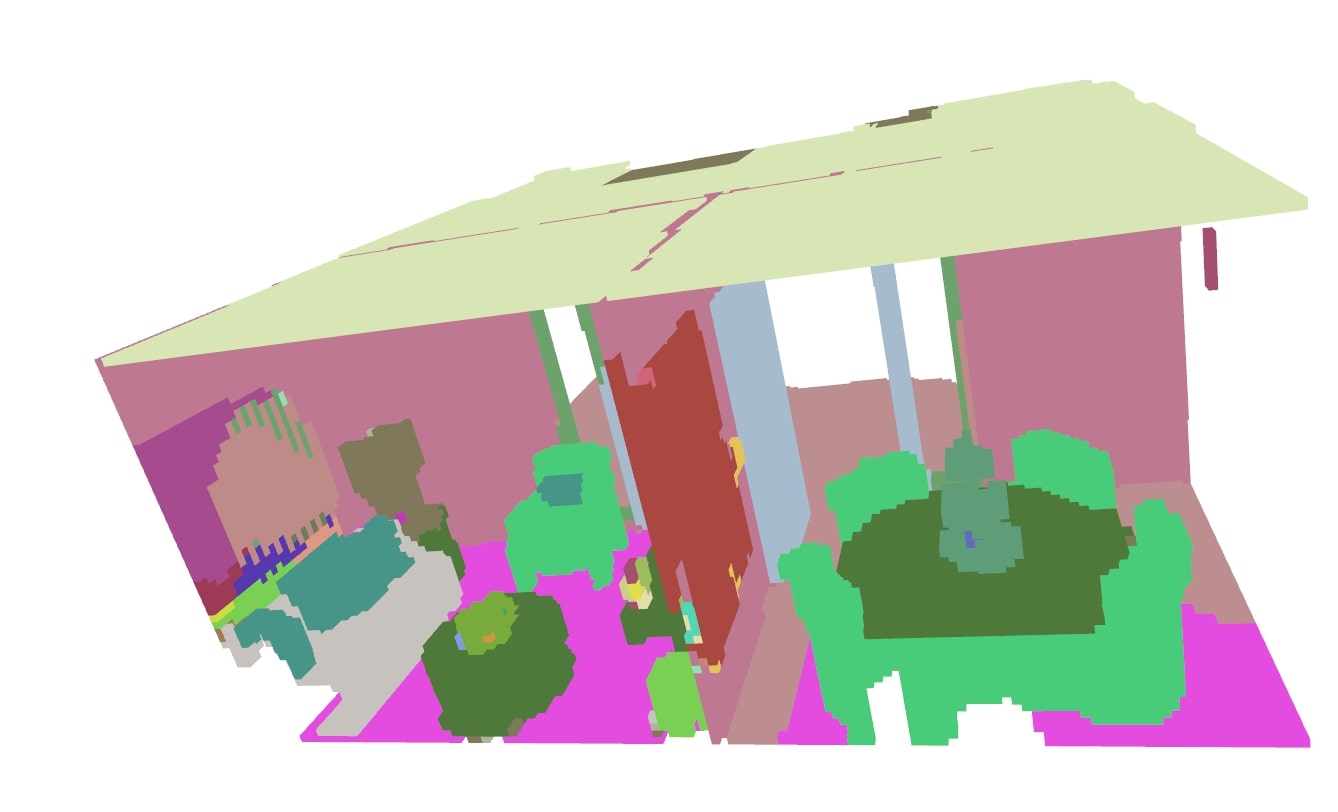}
    \end{subfigure} \\[-2pt]

    \begin{subfigure}[b]{0.46\columnwidth}
        \includegraphics[width=\linewidth,height=0.1\textheight]{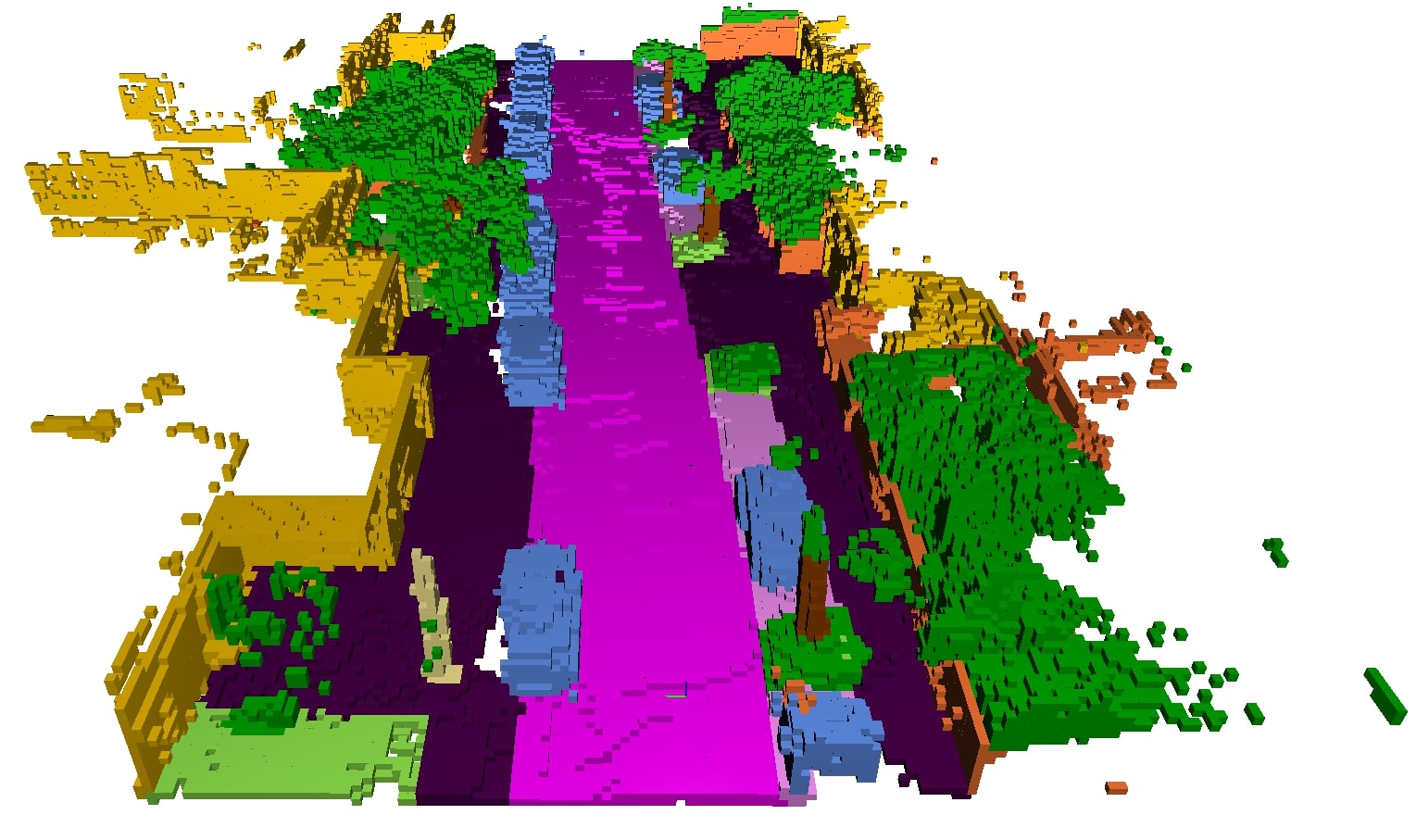}
    \end{subfigure} &
    \begin{subfigure}[b]{0.46\columnwidth}
        \includegraphics[width=\linewidth,height=0.1\textheight]{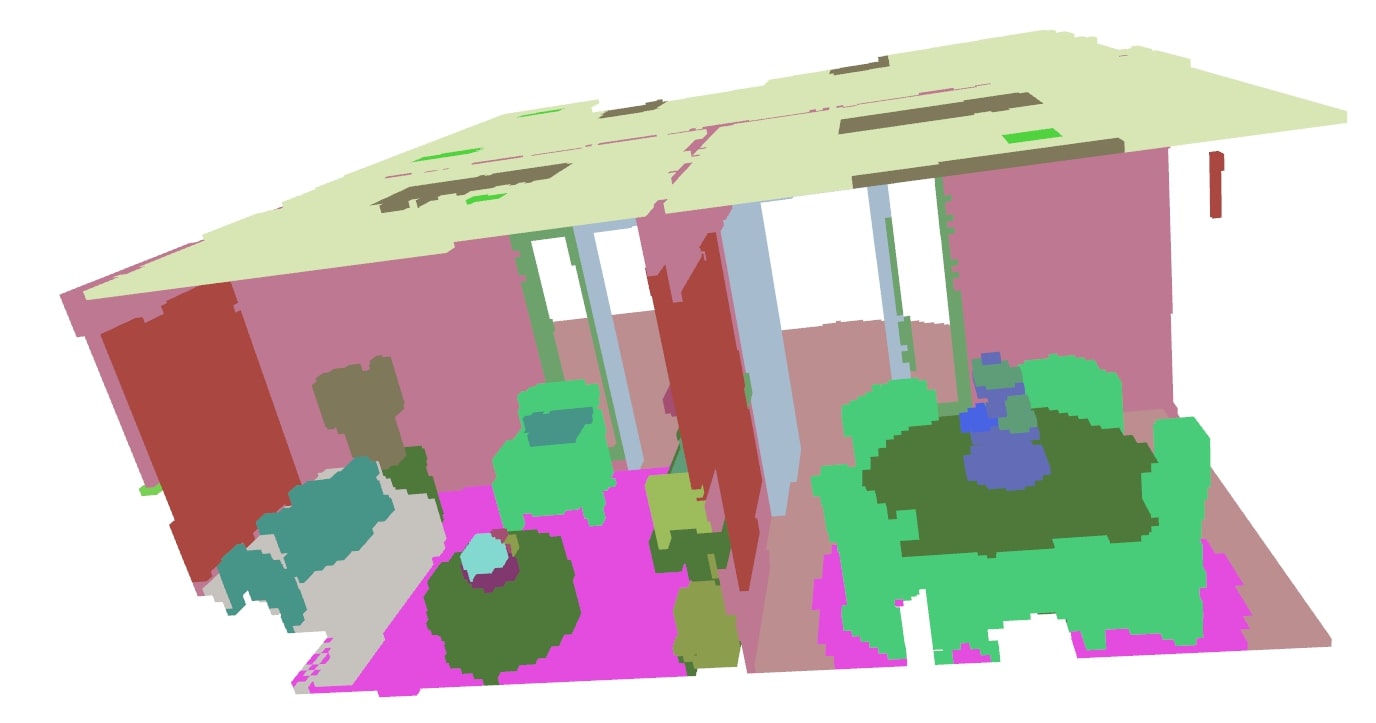}
    \end{subfigure}
\end{tabular}

\caption{\textbf{Latent Semantic Generation.} 
Each column shows an example scene. 
\textbf{Top:} Input downsampled occupancy structure. 
\textbf{Middle:} Semantic scene generated via latent semantic diffusion and VAE decoding, conditioned on the structure. 
\textbf{Bottom:} Ground-truth semantic map. }
\label{fig:latent-semantic-generation}
\end{figure}

\begin{figure}[t]
\centering
\setlength{\tabcolsep}{1pt} 
\renewcommand{\arraystretch}{0.8} 
\captionsetup[subfigure]{labelformat=empty} 

\begin{tabular}{@{}c@{\hskip 2pt}!{\vrule width 0.4pt}@{\hskip 2pt}c@{}}
    \begin{subfigure}[b]{0.46\columnwidth}
        \includegraphics[width=\linewidth,height=0.1\textheight]{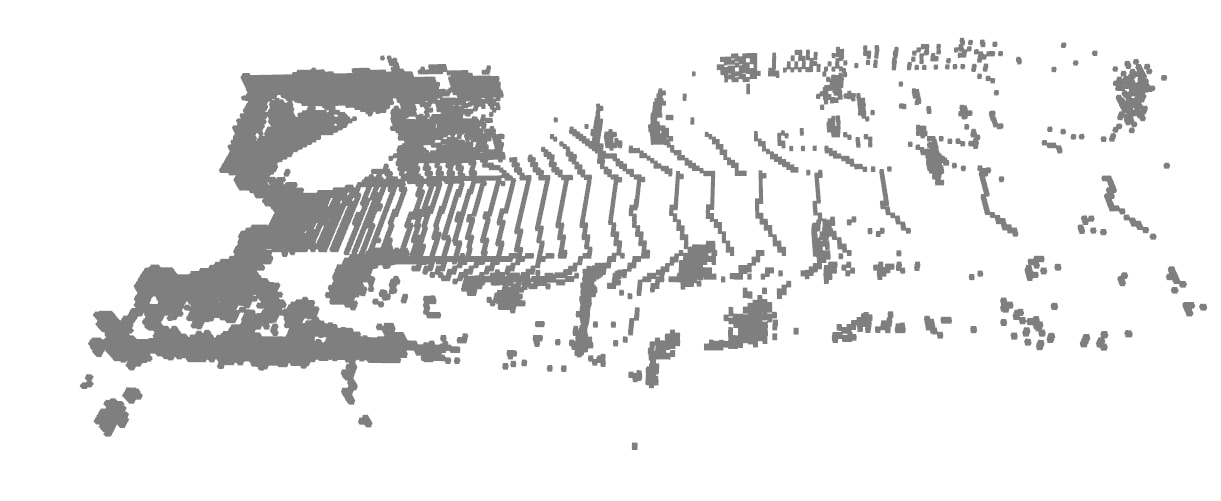}
    \end{subfigure} &
    \begin{subfigure}[b]{0.46\columnwidth}
        \includegraphics[width=\linewidth,height=0.1\textheight]{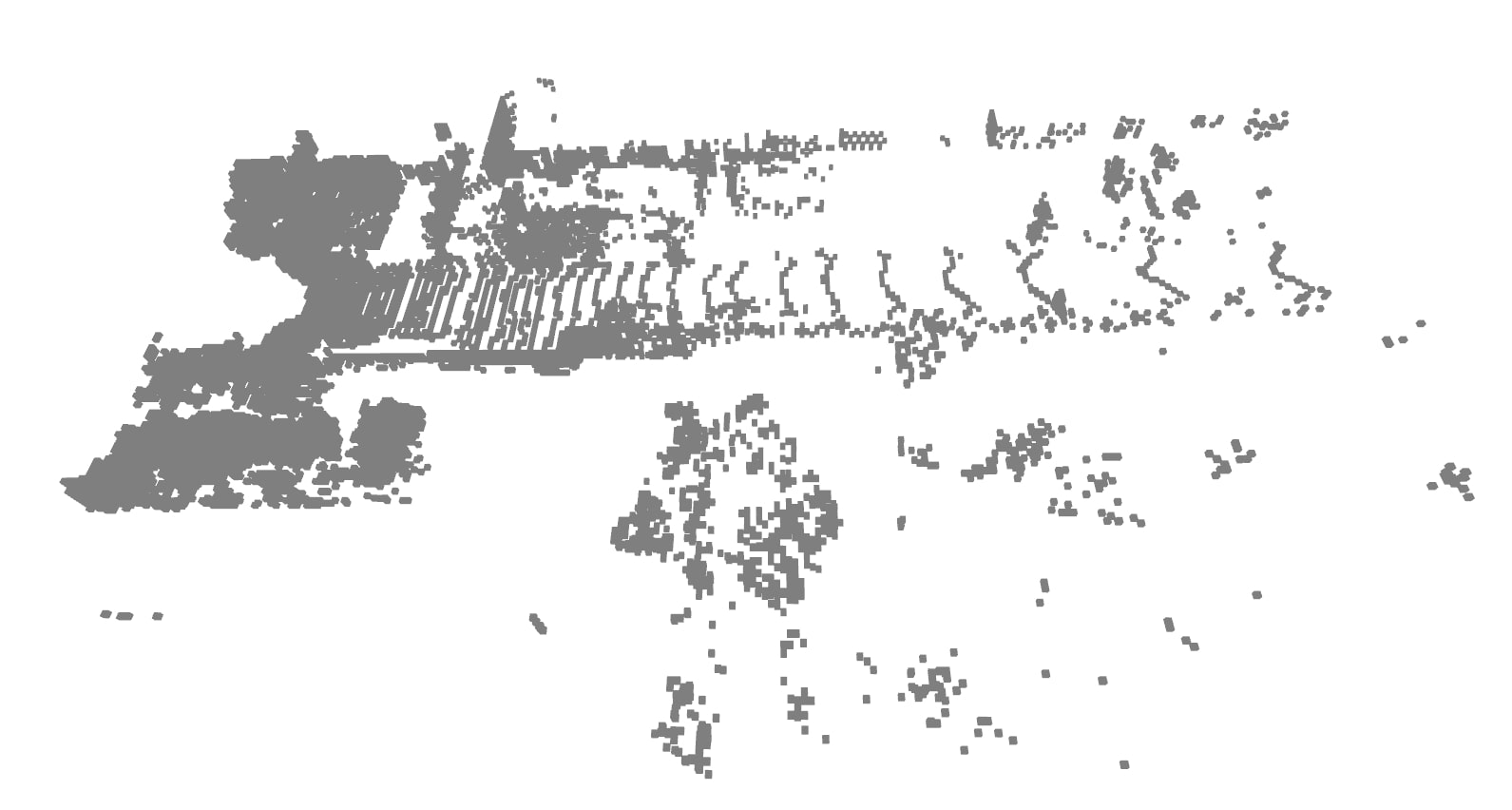}
    \end{subfigure} \\[-2pt]

    \begin{subfigure}[b]{0.46\columnwidth}
        \includegraphics[width=\linewidth,height=0.1\textheight]{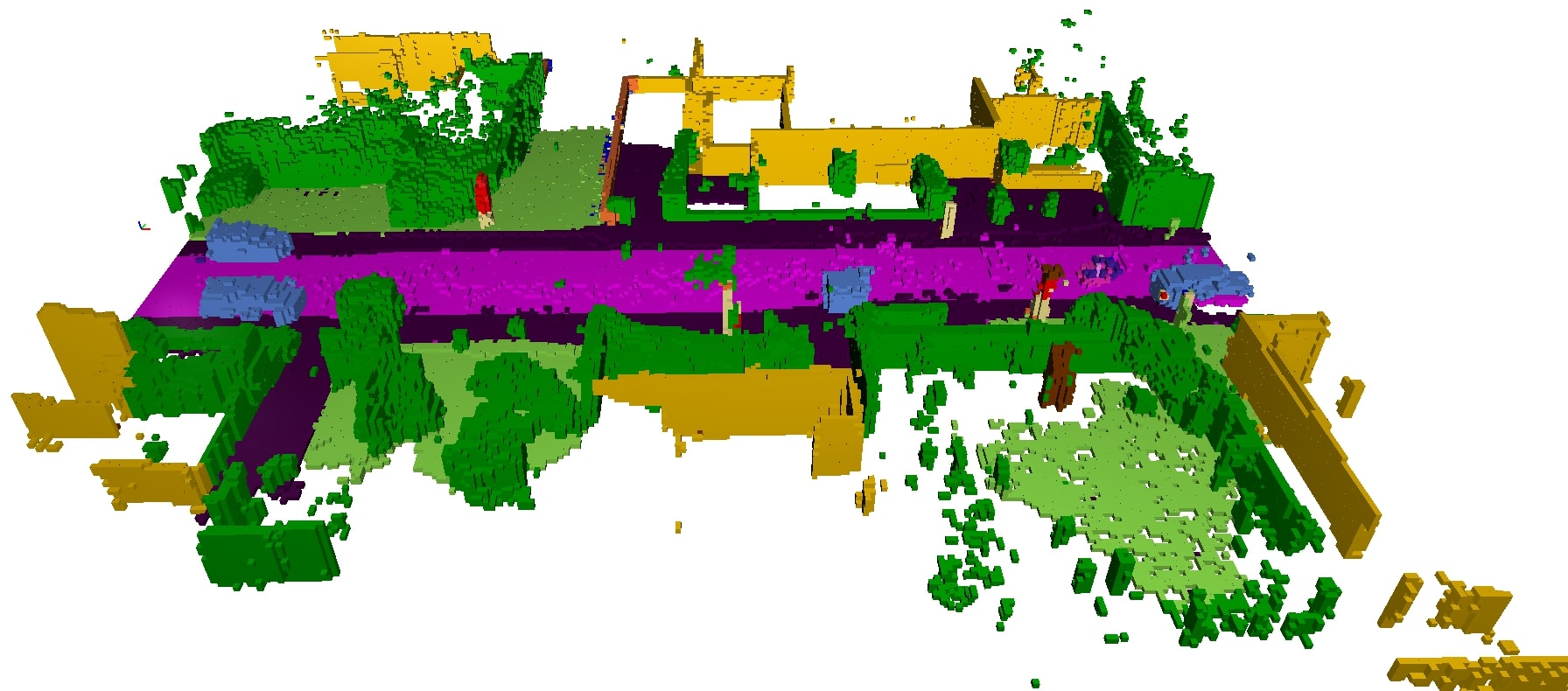}
    \end{subfigure} &
    \begin{subfigure}[b]{0.46\columnwidth}
        \includegraphics[width=\linewidth,height=0.1\textheight]{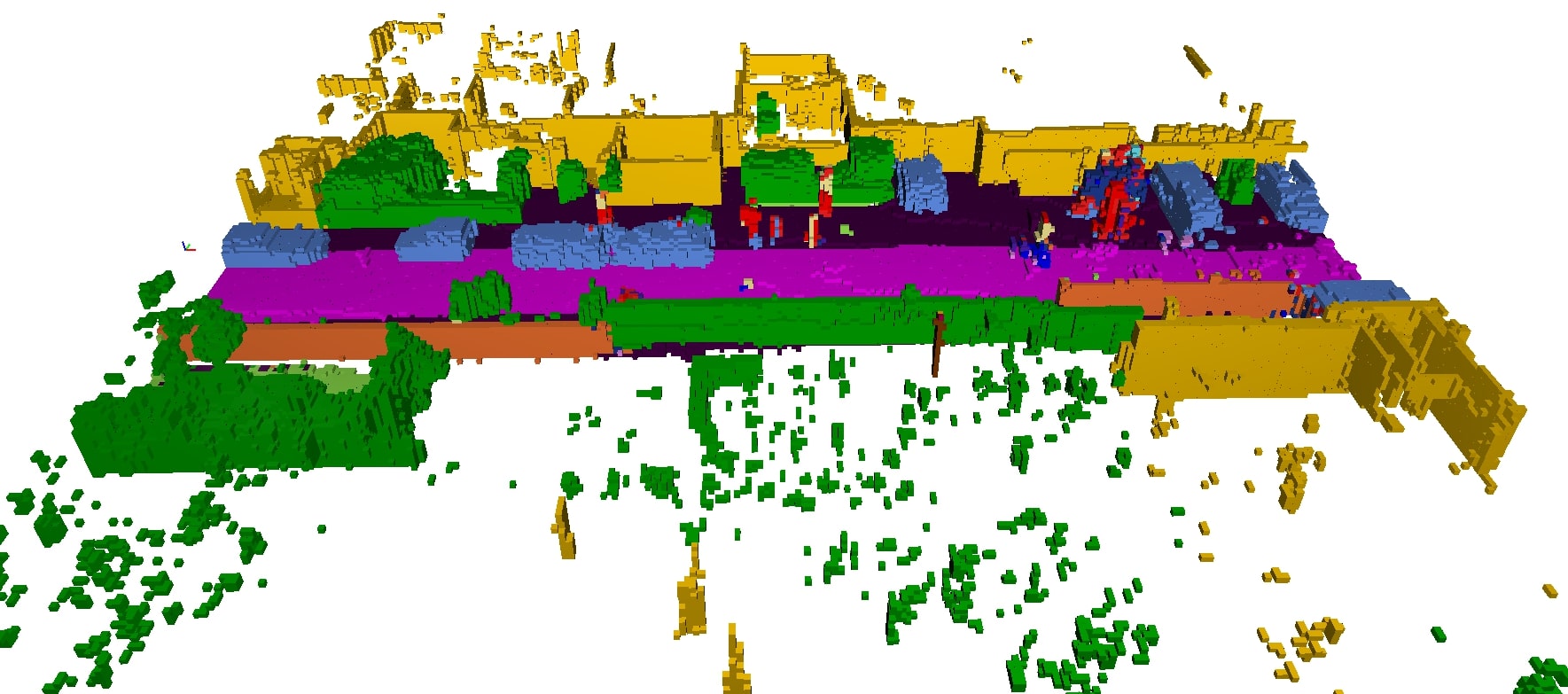}
    \end{subfigure} \\[-2pt]

    \begin{subfigure}[b]{0.46\columnwidth}
        \includegraphics[width=\linewidth,height=0.1\textheight]{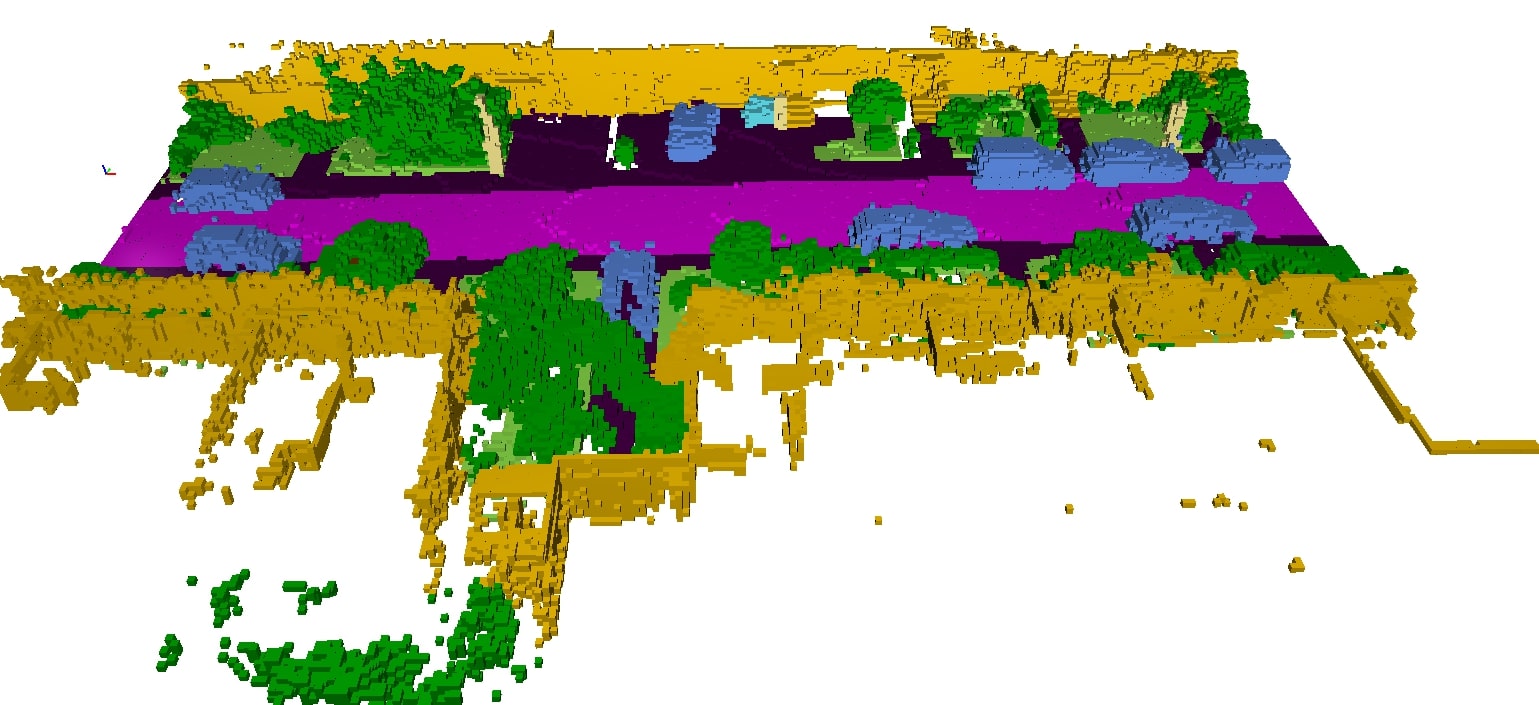}
    \end{subfigure} &
    \begin{subfigure}[b]{0.46\columnwidth}
        \includegraphics[width=\linewidth,height=0.1\textheight]{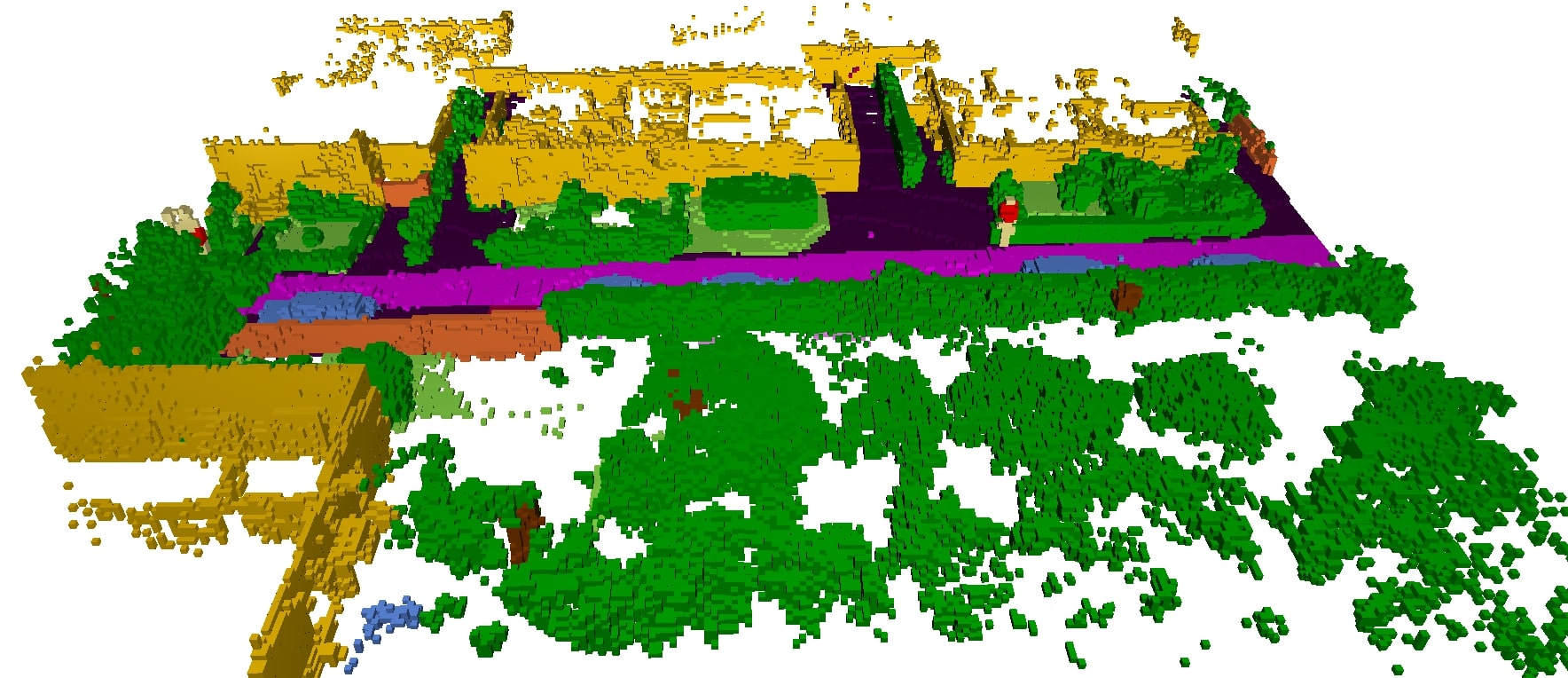}
    \end{subfigure}
\end{tabular}

\caption{{Semantic Scene Completion.} 
Each column shows an example scene. 
\textbf{Top:} Input a single LiDAR scan. 
\textbf{Middle:} Generated semantic map.
\textbf{Bottom:} Ground-truth semantic map. }
\label{fig:semantic-scene-completion}
\end{figure}

\begin{figure}[t]
\centering
\setlength{\tabcolsep}{1pt} 
\renewcommand{\arraystretch}{0.8} 
\captionsetup[subfigure]{labelformat=empty} 

\begin{tabular}{@{}c@{\hskip 2pt}!{\vrule width 0.4pt}@{\hskip 2pt}c@{}}
    \begin{subfigure}[b]{0.46\columnwidth}
        \includegraphics[width=\linewidth,height=0.1\textheight]{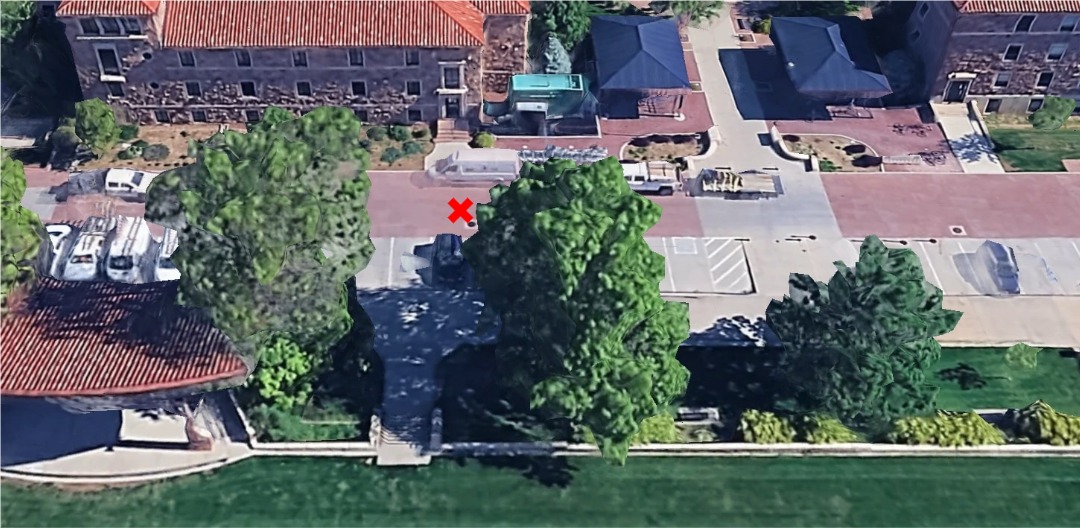}
    \end{subfigure} &
    \begin{subfigure}[b]{0.46\columnwidth}
        \includegraphics[width=\linewidth,height=0.1\textheight]{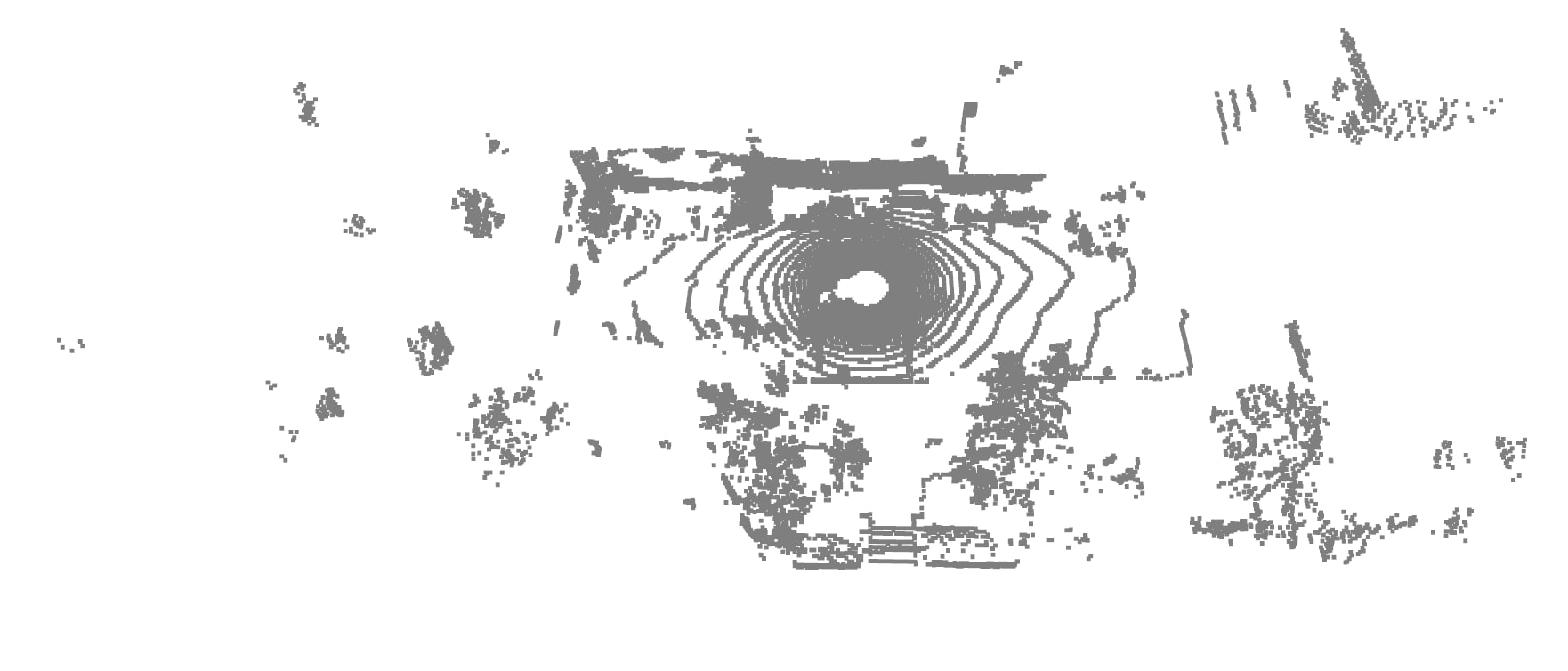}
    \end{subfigure} \\[-2pt]

    \begin{subfigure}[b]{0.46\columnwidth}
        \includegraphics[width=\linewidth,height=0.1\textheight]{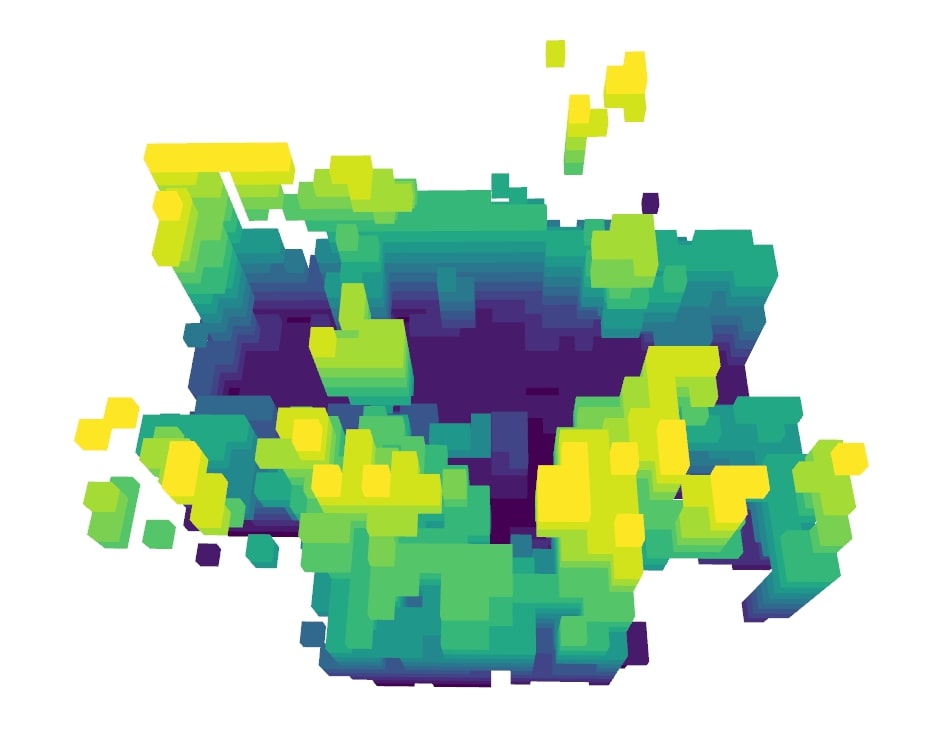}
    \end{subfigure} &
    \begin{subfigure}[b]{0.46\columnwidth}
        \includegraphics[width=\linewidth,height=0.1\textheight]{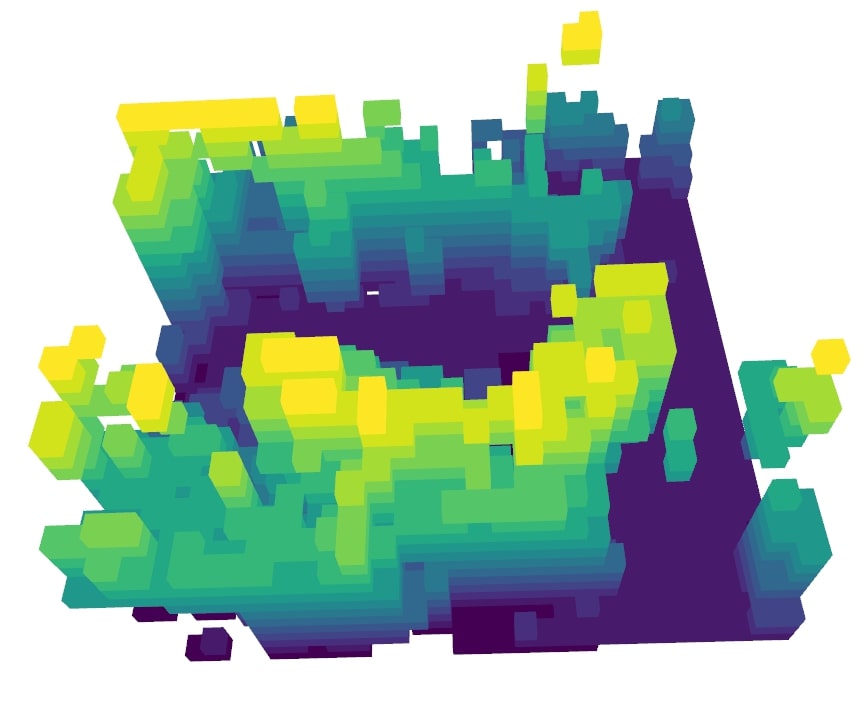}
    \end{subfigure} \\[-2pt]

    \begin{subfigure}[b]{0.46\columnwidth}
        \includegraphics[width=\linewidth,height=0.1\textheight]{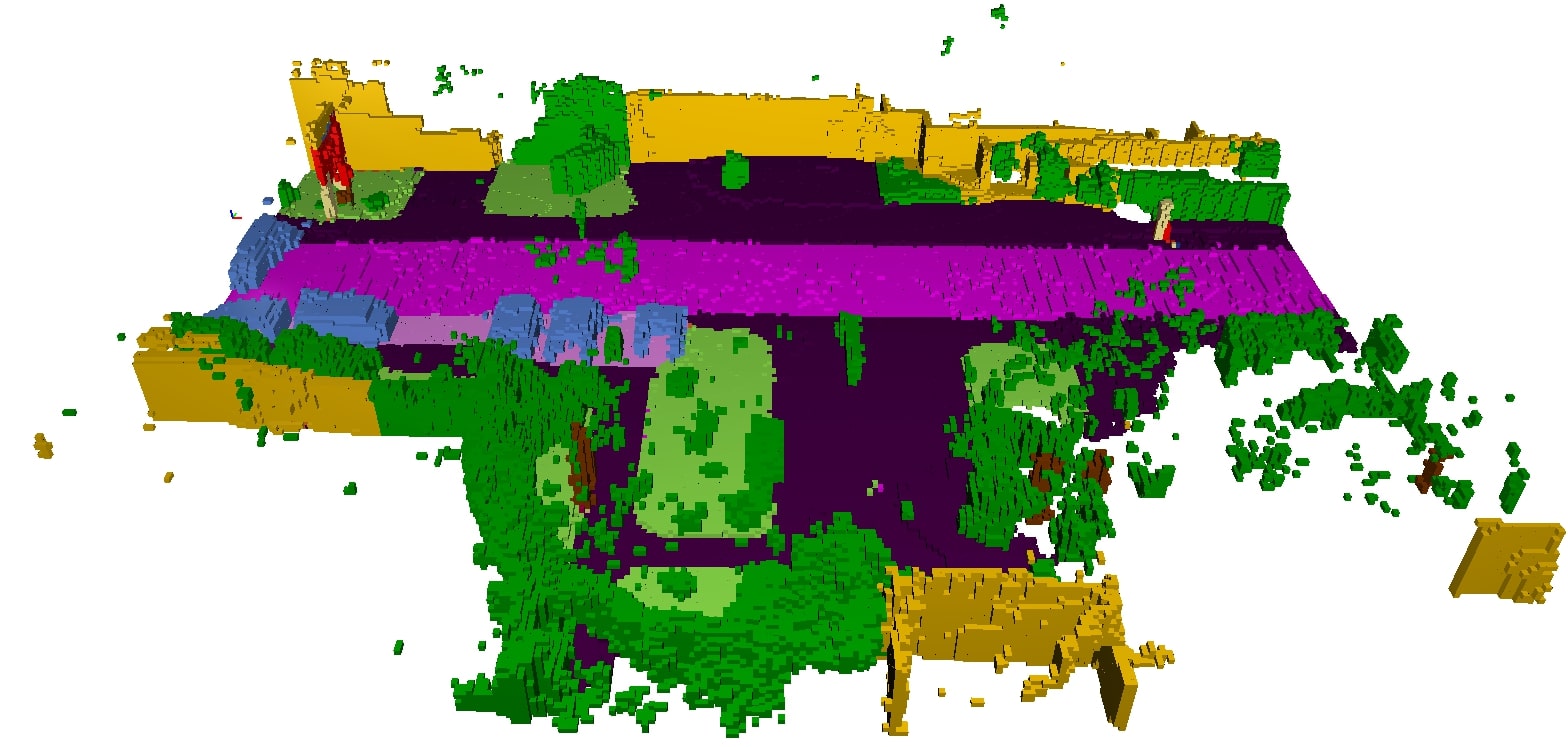}
    \end{subfigure} &
    \begin{subfigure}[b]{0.46\columnwidth}
        \includegraphics[width=\linewidth,height=0.1\textheight]{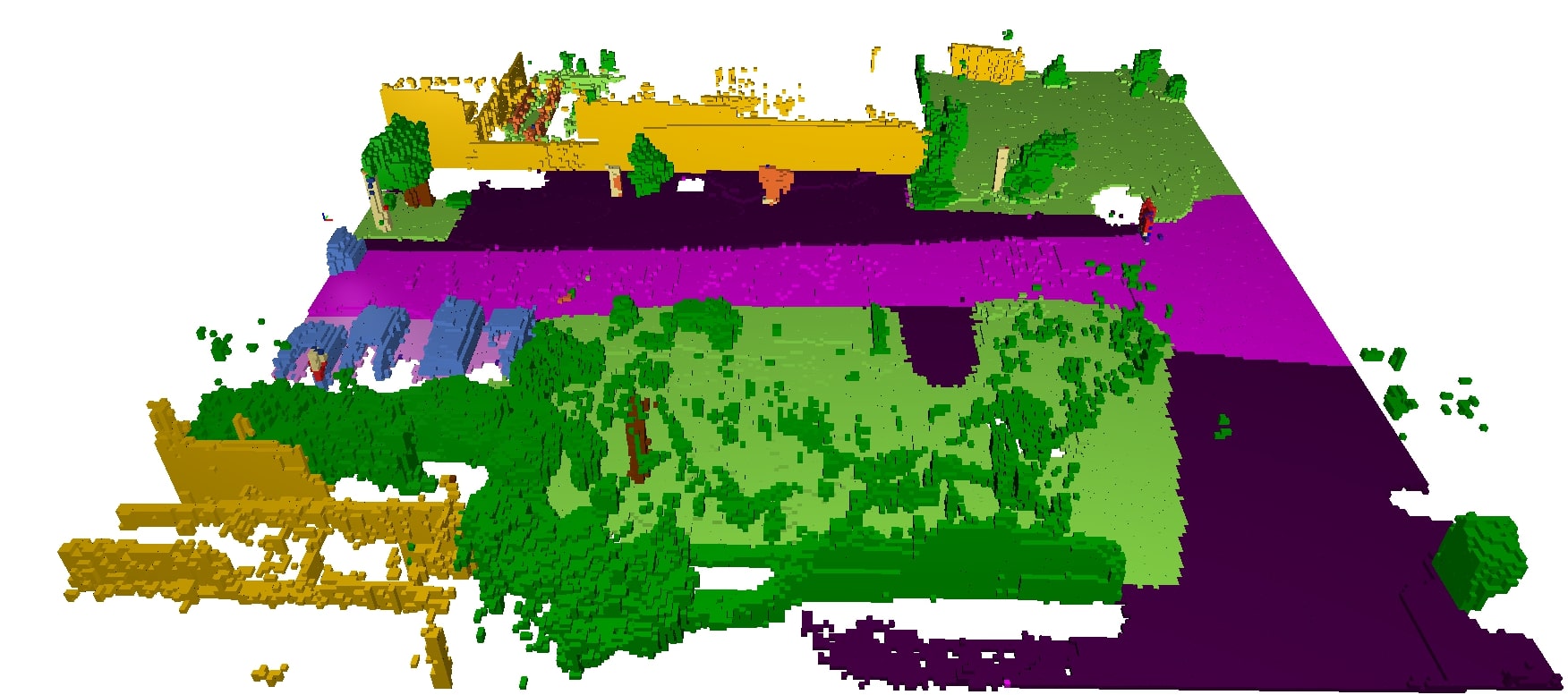}
    \end{subfigure}
\end{tabular}

\caption{{Zero Shot transfer to CU-Multi Dataset.} 
\textbf{Top Left:} satellite view of the scene, where the red crossing denotes the position of the robot.
\textbf{Top Right:} input LiDAR scan from scene via CU-Multi dataset.(Note that LiDAR scan and satellite image is collected on different days.)
\textbf{Middle:} inpainted latent structure of the scene.
\textbf{Bottom:} output semantically labeled scene from two-stage generation with latent structural inpainting. }
\label{fig:zero_shot}
\end{figure}

\begin{figure}[t] 
    \centering
    \begin{subfigure}[b]{0.48\columnwidth}
        \centering
        \includegraphics[width=\linewidth]{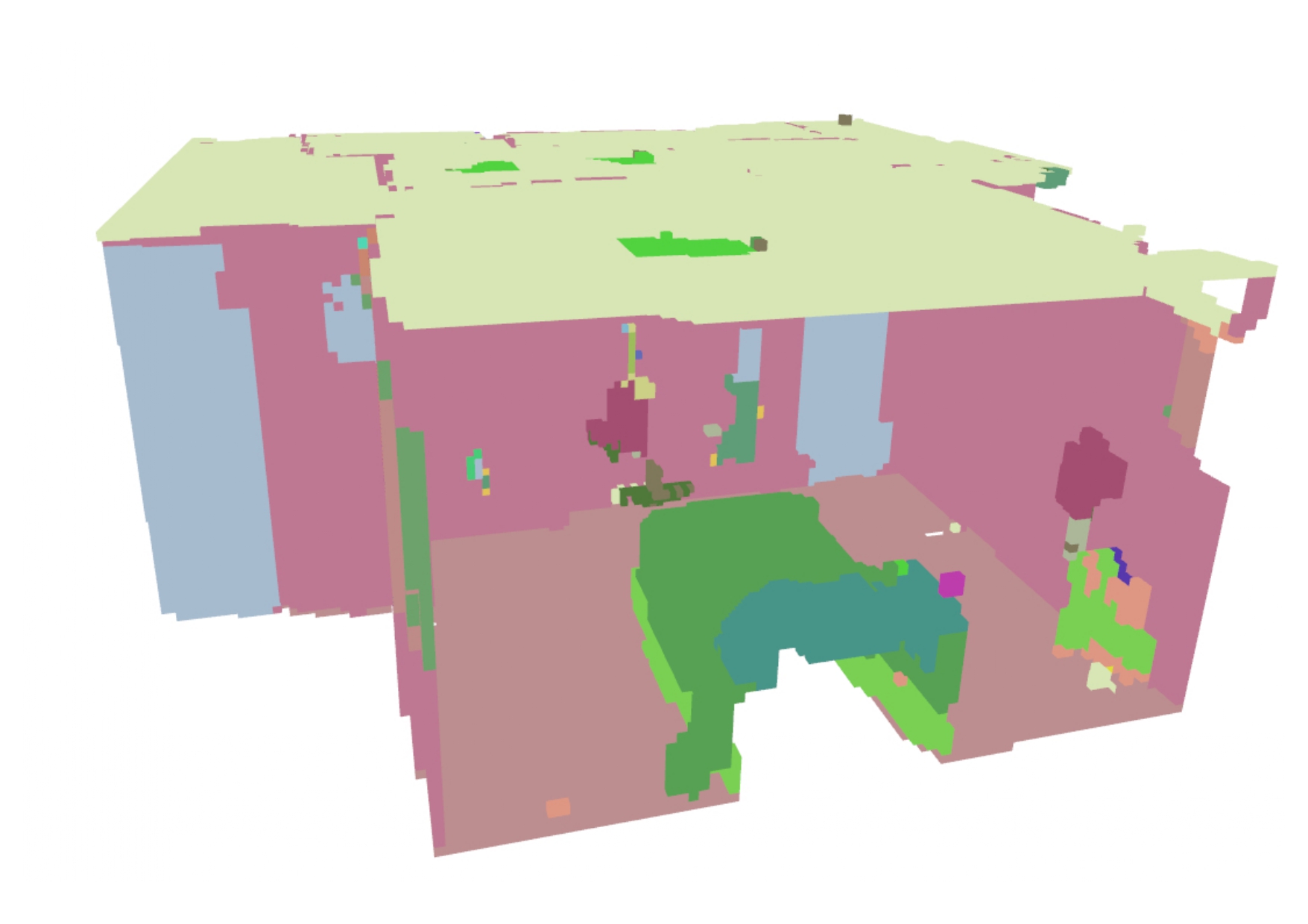}
        
    \end{subfigure}
    \hfill
    \begin{subfigure}[b]{0.48\columnwidth}
        \centering
        \includegraphics[width=\linewidth]{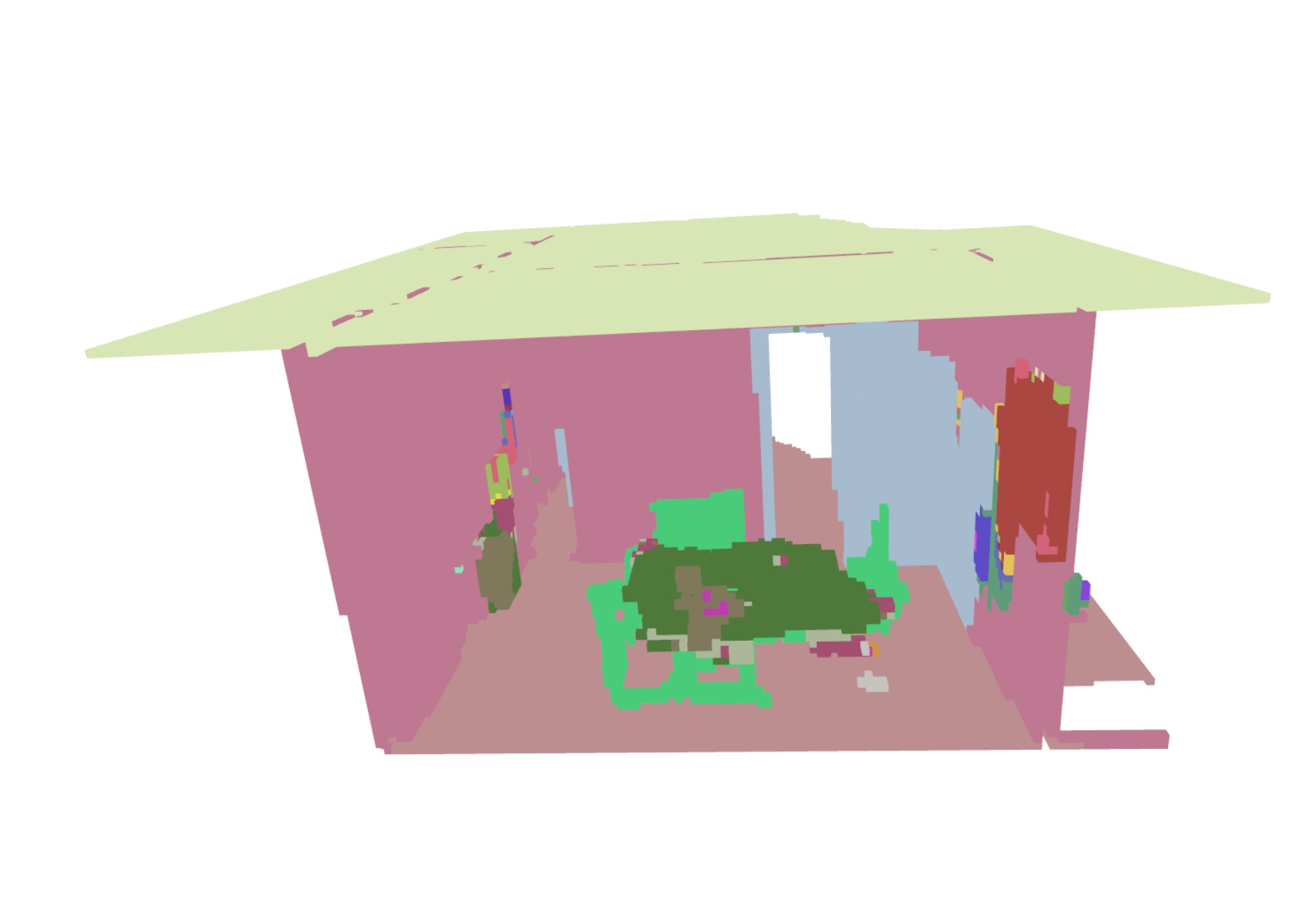}
    
    \end{subfigure}

    \caption{Generations of indoor scenes}
    \label{fig:indoor-scene-generation}
\end{figure}

\begin{table}[t]
\centering
\small 
\setlength{\tabcolsep}{6pt} 
\renewcommand{\arraystretch}{0.95} 

\begin{tabular}{|l|c|c|c|}
\hline
\textbf{Method} & \textbf{FID ↓} & \textbf{KID ↓} & \textbf{MMD ↓} \\ \hline
2-stage Generation (Ours)        & 0.2107 & 8.8700 & 0.1856 \\ \hline
Semantic Generation (Ours) & 0.0289 & 0.7622 & 0.0265 \\ \hline
Semantic Completion (Ours) & 0.0353 & 0.6980 &  0.0216\\ \hline
SemCity (Baseline)        & 0.0940 & 2.2256 &  0.0773\\ \hline
\end{tabular}

\caption{Distributional metrics (FID, KID, and MMD) between generated scenes and the SemanticKITTI validation set. Note that KID is scaled by $\times10^3$. We compare our two-stage generation pipeline, latent-semantic generation, and semantic scene completion against the SemCity baseline.}
\label{tab:outdoor-fid-kid}
\end{table}

\begin{table}[t]
\centering
\small
\setlength{\tabcolsep}{5pt}
\renewcommand{\arraystretch}{1.2}

\begin{tabular}{|l|c|c|c|}
\hline
\textbf{Method} & \textbf{FID ↓} & \textbf{KID ↓} & \textbf{MMD ↓} \\ \hline
2-stage Generation (Ours)        & 0.8671 & 0.5245 & 0.2837 \\ \hline
Semantic Generation (Ours) & 0.0486 & 0.1193 & 0.0668 \\ \hline
SceneSense (Baseline)  & 31.0950 & 89.7884 & 47.8399 \\ \hline
SemCity (Baseline)         & 50.2789 & 187.2540 & 99.5552 \\ \hline
\end{tabular}

\caption{FID, KID, and MMD between pure generation and the Replica validation set. 
Note that FID, KID and MMD values are scaled by $10^{4}$, $10^{5}$ and  $10^{4}$ for respectively. 
We compare our 2-stage generation with triplane- and voxel-based representations \cite{semcity,SceneSense}.}
\label{tab:indoor_fid_kid}
\end{table}





\subsection{Scene Generation}
We first evaluate unconditional scene generation on both outdoor and indoor datasets. 
To quantify distributional similarity between generated and real scenes, we compute three complementary metrics comparing the distribution of generated samples with that of the validation set:

\textbf{Fréchet Inception Distance (FID)} computes the Fréchet distance between the Gaussian approximations of real and generated feature distributions. Lower FID indicates closer alignment in feature space. \textbf{Kernel Inception Distance (KID)} is an unbiased estimate of the squared Maximum Mean Discrepancy (MMD) between Inception features, computed with a degree-3 polynomial kernel. \textbf{Maximum Mean Discrepancy (MMD)} is additionally computed directly on our VAE latent features using an Radial Basis Function (RBF) kernel. This captures distributional differences in the model’s native representation space.

Together, these metrics provide a robust measure of both global distributional alignment and task-specific latent consistency. Specifically, we train a bird's-eye-view VAE to compute metrics in outdoor generation, and train a 3D VAE for indoor generation.  We present these results alongside comparable methods in tables~\ref{tab:outdoor-fid-kid} and \ref{tab:indoor_fid_kid}. Qualitative results (Figures \ref{fig:outdoor-scene-generation} and \ref{fig:indoor-scene-generation}) demonstrate that the generated samples capture both the global structure (roads, buildings, room layouts) and fine-grained semantics (cars, pedestrians, furniture). We modify the voxel-based diffusion model~\cite{SceneSense} and triplane latent diffusion model~\cite{semcity} to compare in indoor semantic generation in table~\ref{tab:indoor_fid_kid}. 

To evaluate the latent semantic diffusion in isolation, we run stage-2 latent semantic diffusion using the ground-truth coarse occupancy as input, bypassing the structure generation stage. Given the ground truth down-sampled structure, the model achieves competitive metrics, indicating its strong capability to generate semantics that remain plausible and consistent with the data distribution.
Since we operate in a $32{\times}32{\times}16$ downsampled occupancy grid (each voxel corresponds to $8{\times}8{\times}2$ neighboring voxels), semantically correct predictions may be penalized in IoU if they shift by several cell in original space.
Figure~\ref{fig:semantic-scene-completion} illustrates this ``mask dilation" effect: while predictions align structurally with the ground truth, they are not perfectly voxel-to-voxel aligned.

\begin{table}[t]
\centering
\small
\begin{tabular}{lcccccc}
\toprule
Class & Road & Sidewalk & Building & Vegetation & mIoU \\
\midrule
IoU (\%) & 67.6 & 37.0 & 28.1 & 33.5 & 16.9 \\
\bottomrule
\end{tabular}
\caption{Latent Semantic Generation on SemanticKITTI Given Ground-Truth Coarse Occupancy.
}
\label{tab:outdoor-semantic-generation}
\end{table}

\begin{table}[t]
\centering
\small
\begin{tabular}{lcccccc}
\toprule
Class & Floor & Wall & Door & Table & mIoU \\
\midrule
IoU (\%) & 98.57 & 97.89 & 72.49 & 71.58 & 32.22\\
\bottomrule
\end{tabular}
\caption{Latent Semantic Generation on Replica Given Ground-Truth Coarse Occupancy.
}
\label{tab:indoor-semantic-generation}
\end{table}

Results in tables~ \ref{tab:outdoor-semantic-generation} and \ref{tab:indoor-semantic-generation} show that our model nonetheless produces coherent semantics across major classes such as road, sidewalk, vegetation, floor and wall. Given ground-truth coarse occupancy in both SemanticKITTI and Replica validation dataset, we generate semantics via latent diffusion. Results show strong performance on major scene classes relevant for navigation and planning.
\subsection{Semantic Scene Completion}
Next, we test semantic scene completion from sparse partial LiDAR scans. We downsample a single input scan and project it into the voxelized structure space, then apply structure diffusion to inpaint the missing geometry. Semantic diffusion is then applied on top of the completed structure to recover semantic occupancy. As shown in Figure~\ref{fig:semantic-scene-completion} and table~\ref{tab:outdoor-fid-kid}, our blended diffusion produces coherent completions that respect local context while maintaining global consistency. Conditioned on a single LiDAR scan, our model generates scenes of comparable quality to those produced when conditioned on the complete oracle structure. Outpainting further extends the scene beyond the observed region, producing plausible continuations (Figure \ref{fig:outdoor-scene-generation}).
\subsection{Zero-Shot Generalization}
To assess the generalization ability of our model, we apply our trained structure and latent semantic diffusion models to new domains without any retraining or fine-tuning.
We test on LiDAR scans from CU-Multi~\cite{albin2025cumultidatasetmultirobotdata}, a dataset collected in outdoor areas of a campus, which differs from SemanticKITTI in LiDAR configuration (e.g. height and tilt angle) and scene layout.  
Given a single LiDAR scan from the dataset, our model successfully completes missing structure and generates a series of plausible semantic maps aligned with real-world layouts, despite the domain shift (Figure~\ref{fig:semantic-scene-completion}). This highlights the robustness of our dual octree graph latent diffusion framework and its potential for real-world deployment.

\section{Limitations and Future Work}
Although our two-stage framework successfully disentangles structure generation from latent semantic diffusion, we observe that the overall generation quality is still bottlenecked by the structure generation stage. 
In our current implementation, structure generation is performed with 3D CNNs at a coarse occupancy level. 
While more advanced or domain-specific models could improve this stage, generating plausible structures without semantic guidance remains a challenging problem.

Another limitation arises from mask handling during post-conditioned generation. 
Because masks must be defined in the downsampled latent space, a single masked voxel corresponds to a larger region in the original space. 
This leads to slight mask dilation when mapped back to the high-resolution voxel grid, causing imperfect alignment between the conditioned input and the final generation, which is reported as a common limitation for latent-space postconditioning methods~\cite{LatentInpaint}.

For future work, we plan to explore classifier-free guidance to more explicitly unify indoor and outdoor scene generation within a single model. 
A major challenge is that existing datasets are typically domain-specific and define different semantic taxonomies, making joint training difficult. 
One promising direction is to adopt open-vocabulary semantic representations, allowing cross-domain training without label-space conflicts. 
Finally, while this work is evaluated offline, our method is efficient enough to be deployed in online settings. 
We plan to integrate it into indoor and outdoor robotic navigation systems to enable real-time semantic completion and scene reasoning.

\section{Conclusion}
We introduced a dual octree graph latent diffusion framework that is capable of generating, completing, and extending 3D semantic scenes in a single architecture. Experiments on SemanticKITTI, Replica, and zero-shot inference highlight its flexibility, and generation quality. Overall, our results suggest that completion-through-generation in a locality-preserving structured latent space is a viable direction for robotic perception.

\section*{Acknowledgment}
The authors acknowledge the use of GPT-4o (OpenAI) for refining the literature review in Section II and methods in Section III. The generated content was reviewed and edited by the authors to ensure technical accuracy and consistency with the overall manuscript.
\printbibliography

\end{document}